\PassOptionsToPackage{table}{xcolor}
\documentclass[]{academic_template}
\usepackage[toc,page,header]{appendix}

\usepackage{marvosym}

\usepackage{bbding}
\usepackage{soul}

\usepackage{amssymb}
\usepackage{amsmath}

\usepackage{makecell}
\usepackage{algorithm}
\usepackage{algpseudocode}
\usepackage{tabularx}
\usepackage{booktabs}
\usepackage{graphicx}
\usepackage{array}
\newcolumntype{C}{w{c}{3em}}
\usepackage{listings}
\usepackage{hyperref}
\usepackage{url}
\usepackage{booktabs}
\usepackage{wrapfig}

\usepackage{xcolor}

\usepackage{colortbl}
\usepackage{caption}
\usepackage{subcaption}

\usepackage{amsthm}

\setlogoheight{8mm}

\setlogospacing{0mm}
\setsjtublue

\settitlerulethickness{3pt}

\abstractboxon
\setabstractframecolor{gray}
\setabstractbgcolor{gray!10}
\setlogotolineshift{0mm}

\settoprulethickness{2.5pt}
\setbottomrulethickness{1.5pt}

\usepackage{dblfloatfix}
\usepackage{float}

\usepackage{tabularx}

\usepackage{multicol}
\usepackage{multirow}
\usepackage[table]{xcolor}

\usepackage{natbib}
\setcitestyle{numbers}
\setcitestyle{square}
\usepackage{graphicx}

\usepackage{xspace}

\newcommand{\ours}{ACE-Brain-0.7\xspace}
\newcommand{\oursTwoLine}{\makecell{ACE-Brain-\\0.7}}

\newcommand{\base}{Qwen3.5-9B\xspace}
\newcommand{\astra}{GPT-6 Astra\xspace}
\newcommand{\opus}{Claude Opus~5.5\xspace}

\title{ACE-Brain-0.7: Learning Execution-Grounded \\
Reasoning for Embodied Agents}

\author{%
\begin{minipage}{\textwidth}
    \centering
    \sffamily
    \setlength{\parskip}{0pt}

    {\fontsize{12}{16}\selectfont
    ACE-Brain Team\par}

    \vspace{1.2em}

    {\fontsize{10}{14}\selectfont
    Please see
    \hyperref[app:contributors]{Contributions and Author List}
    for more author details.\par}
\end{minipage}%
}

\abstract{
Reliable reasoning for long-horizon embodied tasks requires predictions grounded in the observed execution state.
We present \ours, a 9B vision--language model that combines spatial grounding, task understanding, and progress estimation.
To train the model, we curate 3.7 million question--answer pairs, drawing execution supervision from egocentric human videos, robot episodes, and simulation.
These questions span different stages of an execution and connect task-level judgments with observed progress and outcomes.
We then fine-tune \base on this collection together with additional task, spatial-grounding, and progress-estimation data.
Across six public task-reasoning benchmarks, \ours achieves an average score of 76.5, ahead of \astra at 72.7 and \opus at 73.5.
Our model also leads the evaluated models in spatial grounding, averaging 75.90 across seven benchmarks and reaching 91.25 Acc@0.5 on RefCOCOg.
Controlled evaluations further indicate that our task-only model makes more effective use of visual history and more accurately distinguishes execution states.
Together, these results show that \ours effectively combines spatial perception with temporal understanding, delivering strong execution-grounded reasoning and comparing favorably with closed-source models.
}

\date{\today}

\newif\ifusebibliography
\usebibliographyfalse

\begin{document}

\phantomsection
\addcontentsline{toc}{section}{Abstract}
\maketitle

\newpage
\tableofcontents

\newpage
\section{Introduction}
\label{sec:intro}

\begin{figure}[!t]
    \centering
    \includegraphics[width=\linewidth]{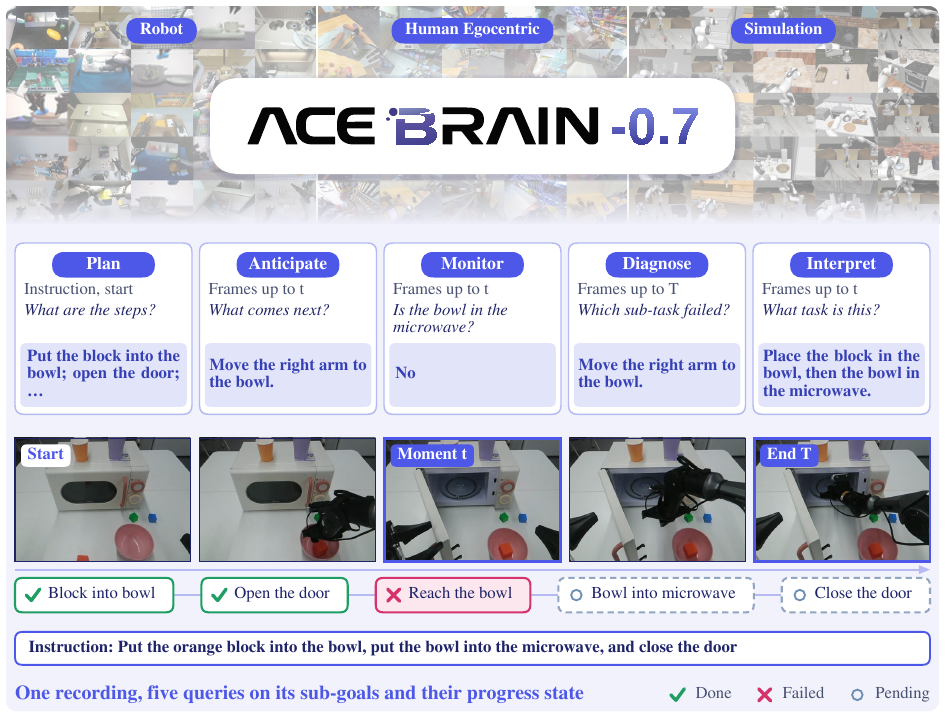}
    \caption{
        \textbf{Overview of \ours.}
        \emph{Top:} question--answer pairs derived from annotated robot episodes, egocentric human video and simulation fine-tune \base into \ours.
        \emph{Bottom:} one failed execution of a robot task, with its sub-goals (abridged) and their state at the end of the recording (\emph{pending}: not started). The five capabilities are five queries on these sub-goals and their state; $t$ and $T$ mark the frames a query sees.
    }
    \label{fig:overview}
\end{figure}

Recent advances in embodied AI have equipped robots with a growing range of manipulation skills.
Completing complex, long-horizon tasks requires high-level reasoning that combines task goals with environmental observations to coordinate these skills, assess execution state, and select the next action.
Systems such as Hi Robot~\cite{shi2025hirobot}, $\pi_{0.7}$~\cite{pi07}, and Gemini Robotics~\cite{geminirobotics15_2025} demonstrate how vision--language models (VLMs) can support task planning, sub-task prediction, and execution supervision.
However, a plausible plan does not guarantee an accurate judgment of execution state.
Task knowledge tells a model which steps are usually required, but deciding when to advance requires observations of what has actually happened.
In Figure~\ref{fig:overview}, the robot attempts to place the block in the bowl and opens the microwave door but fails to reach the bowl; at this state, issuing a placement or door-closing command would skip an unresolved step. 
Recognizing this situation requires the model to relate the observed scene to the task's progress and completion conditions.

We use \textbf{\emph{execution-grounded reasoning}} to describe judgments that follow the observed progress and outcome of a task.
This motivates supervision that covers different stages of execution, including incomplete and failed attempts.
To learn this ability, we curate a training collection of 3.7 million question--answer pairs centered on task execution.
Specifically, we collect supervision from egocentric human videos, robot episodes, and simulation, with questions posed at multiple moments and answers derived from step timelines, object states, and failure labels.
For newly annotated egocentric videos, we also retain the frames supporting each judgment.
We combine this collection with additional task, spatial-grounding, and progress-estimation data to fine-tune \base~\cite{qwen3.5}, obtaining \ours, a 9B VLM that supports all three capabilities.

\ours outperforms every evaluated open model on each of the six public task-reasoning benchmarks.
It also achieves the highest average scores among the evaluated models on both benchmark suites: 76.5 for task reasoning and 75.90 across seven spatial-grounding benchmarks.
On RoboBench~\cite{luo2025robobench}, \ours scores 52.1 on next-subtask prediction, exceeding 34.7 for \astra~\cite{openai2026gpt6astra} and 39.0 for \opus~\cite{anthropic2026opus55}.
To examine the contribution of the question--answer collection, we also compare \base with a task-only variant fine-tuned on this collection alone.
Experiments with controlled visual inputs indicate that the trained model benefits more from ordered visual history, while temporal question pairs test its judgments across different execution states.
On 114 pairs, the task-only model raises paired accuracy from 15.8 for \base to 54.4.

Our contributions are threefold:
\begin{itemize}\setlength{\itemsep}{0pt}\setlength{\parskip}{0pt}
    \item A training collection of 3.7 million question--answer pairs and a data construction pipeline that derives execution-related supervision from human videos, robot episodes, and simulation, with egocentric video serving as a major data source.
    \item \ours, a 9B VLM that jointly supports spatial grounding, task understanding, and progress estimation, achieving leading overall performance across the corresponding evaluations.
    \item An empirical analysis of planning and execution judgments through capability-level comparisons, controlled visual inputs, temporal question pairs, and an ablation of egocentric supervision.
\end{itemize}
\section{Related Work}
\label{sec:related}

\par\medskip
\noindent\textbf{Embodied reasoning models.}\quad
Hierarchical robot systems use vision--language models to interpret task instructions and observations and issue subtask commands to a low-level controller~\cite{shi2025hirobot,geminirobotics15_2025}.
In $\pi_{0.7}$~\cite{pi07}, subtask instructions, subgoal images, and episode metadata provide richer conditioning for policies trained on diverse robot and non-robot data.
Recent embodied foundation models also bring several reasoning capabilities into a single model.
RoboBrain 2.5~\cite{robobrain2_5} combines spatial reasoning with dense progress estimation, while Embodied-R1.5~\cite{yuan2026} integrates planning, grounding, and correction.
PhysBrain 1.5~\cite{physbrain15} jointly models embodied understanding, actions, and future states.
Within this broader effort, our work focuses on learning task judgments from execution observations and examining how those judgments depend on the evidence provided.

\par\medskip
\noindent\textbf{Spatial grounding.}\quad
Spatial grounding connects language descriptions to objects and locations in a scene, often requiring an understanding of spatial relations and placement constraints.
RoboSpatial~\cite{robospatial} provides supervision for these judgments across egocentric, world-centric, and object-centric reference frames.
To support multi-step spatial referring, RoboRefer~\cite{zhou2025roborefer} combines depth-aware perception with spatial reasoning and introduces the RefSpatial dataset and RefSpatial-Bench.
ACE-Brain-0.6~\cite{wang2026acebrain06} scales verifiable grounding supervision from diverse visual sources to improve spatial perception in an open 9B model.
We retain spatial grounding as a core capability, training and evaluating it alongside task understanding and progress estimation.

\par\medskip
\noindent\textbf{Progress estimation and failure analysis.}\quad
Progress estimation and failure analysis provide complementary information about task execution.
GVL~\cite{ma2025gvl} estimates progress by using a pretrained VLM to order shuffled video frames, whereas ProgressLM~\cite{zhang2026progresslm} learns to estimate progress from a task demonstration and a current observation.
Reward learning offers another source of execution feedback: RoboReward~\cite{lee2026roboreward} augments successful robot episodes through instruction relabeling and temporal truncation to represent unsuccessful and partially completed tasks.
More detailed supervision comes from failure annotations.
AHA and Guardian~\cite{duan2025aha,pacaud2025guardian} synthesize execution failures for error recognition and explanation, while RoboFAC~\cite{ye2025robofac} covers task understanding, failure analysis, and correction through questions about simulated and real executions.
These studies inform our use of progress and outcome supervision, including execution states that successful demonstrations alone do not capture.

\par\medskip
\noindent\textbf{Task-understanding data and benchmarks.}\quad
Human and robot videos provide complementary sources of supervision for task understanding.
EPIC-KITCHENS, Ego4D, and HoloAssist~\cite{damen2020epic,grauman2021ego4d,wang2023holoassist} capture human activities and interactions, while DROID and Open X-Embodiment~\cite{khazatsky2024droid,oxe2023} provide robot trajectories with task instructions.
RoboVQA~\cite{robovqa} derives questions about planning, success, and past and future actions from temporally segmented human and robot videos, and demonstrates the benefits of joint training and video conditioning.
Public benchmarks assess these abilities at different levels.
EgoPlan-Bench~\cite{egoplanbench} tests next-action selection from task goals and observed progress; RoboBench and ERIQ~\cite{luo2025robobench,liu2025eriq} cover multiple dimensions of embodied reasoning.
LADDER~\cite{petkov2026ladder} separately examines task decomposition, state tracking, completion verification, and next-subgoal selection.
Building on these data and evaluation settings, our construction emphasizes supervision at different moments of an execution. We complement benchmark scores with controlled visual inputs and temporal question pairs to test whether answers track changes in the observed execution state.
\section{Task Structure, Execution State and Derived Queries}
\label{sec:capabilities}

Consider one execution of a task: an instruction $\ell$ and the observations $o_{1:t}$ up to time $t$. The \emph{structure} of the task is
\begin{equation}
  G = (V, \prec, \Phi), \qquad V = \{g_1, \dots, g_K\},
\end{equation}
where $V$ is the set of sub-goals, $g_i \prec g_j$ means that $g_i$ must hold before $g_j$ can be attempted, and $\Phi = \{\phi_k\}$ gives each sub-goal a completion condition; benchmarks call a sub-goal a step or a sub-task. The \emph{state} of the execution at time $t$ is
\begin{equation}
  S_t = \{(s_t^k, E_t^k)\}_{k=1}^{K}, \qquad s_t^k \in \{\text{\textsc{not started}},\ \text{\textsc{in progress}},\ \text{\textsc{done}},\ \text{\textsc{failed}}\},
  \label{eq:state}
\end{equation}
where the evidence $E_t^k \subseteq \{1, \dots, t\}$ indexes the observations that establish the status, such as the frames in which $\phi_k$ becomes true. $G$ is shared by all executions of $\ell$ and can be inferred from prior knowledge; $S_t$ belongs to one execution and is not determined by $\ell$.

\begin{table}[!htb]
\centering
\caption{The five capabilities as queries on the task structure $G$ and the execution state $S_t$ (Eq.~\ref{eq:state}), with the query each poses on the recording of Figure~\ref{fig:overview}; $T$ is the end of the recording.}
\label{tab:capabilities}
\footnotesize
\setlength{\tabcolsep}{4pt}
\begin{tabular}{@{}>{\raggedright\arraybackslash}p{1.7cm}>{\raggedright\arraybackslash}p{4.2cm}>{\raggedright\arraybackslash}p{1.2cm}>{\raggedright\arraybackslash}p{2.9cm}>{\raggedright\arraybackslash}p{4.4cm}@{}}
\toprule
Capability & Query on $(G, S_t)$ & Reads & Answer derived from & In Figure~\ref{fig:overview} \\
\midrule
\textbf{Plan} & A linear extension of $(V, \prec)$ & $G$ & Annotated sub-task sequence & What are the steps? \emph{Put the block into the bowl; open the door; \dots} \\
\textbf{Monitor} & $s_t^k$ for a named $k$; object state & $S_t$ & Time-stamped steps; state labels & Is the bowl in the microwave? \emph{No.} \\
\textbf{Anticipate} & A member of the frontier $F_t$ & $G$, $S_t$ & Step that actually followed & What comes next? \emph{Move the right arm to the bowl.} \\
\textbf{Diagnose} & $\{k : s_T^k = \text{\textsc{failed}}\}$, with type and location & $S_T$ & Failure annotations & Which sub-task failed? \emph{Move the right arm to the bowl.} \\
\textbf{Interpret} & $\ell$ and scene facts, from the observed states & Inverse & Task and scene labels & What task is this? \emph{Place the block in the bowl, then the bowl in the microwave.} \\
\bottomrule
\end{tabular}
\end{table}

A question about a task is a function of this pair, $a = q(G, S_t)$: a \emph{structure query} depends on $G$ only, a \emph{state query} on $S_t$. The five capabilities are five such queries (Table~\ref{tab:capabilities}); \emph{interpretation} is the inverse query, recovering the task and scene facts from the observed states. \emph{Anticipation} reads both $G$ and $S_t$: it returns a sub-goal on the frontier
\begin{equation}
  F_t = \{k : s_t^k \neq \text{\textsc{done}},\ s_t^j = \text{\textsc{done}} \text{ for all } g_j \prec g_k\},
\end{equation}
which for a totally ordered task without an unresolved failure is the first incomplete sub-goal. Anticipation is thus planning composed with monitoring: when the candidate steps come from the same recording, the question supplies the structure, and a wrong answer misreads the state.

\par\medskip
\noindent\textbf{Execution-grounded reasoning.}\quad
A predictor is \emph{execution-grounded} on a query if, for two observation sequences under one instruction whose states, and hence answers, differ, it returns the answer of each. We measure execution grounding with four input controls, \emph{blind} (no images), \emph{first frame} only, \emph{last frame} only and \emph{shuffled} frame order, and with \emph{temporal question pairs}: over pairs of observations from one recording with the same question and options but different answers, the share of pairs with both answers correct. Two predictions follow, tested in Section~\ref{sec:exp:grounding}: when the observations are withheld, structure queries lose little, state queries lose much for an execution-grounded model, and a model that answers from the task prior loses little on either; and when the question supplies the structure, wrong anticipation answers concentrate on sub-goals adjacent to the frontier.

\section{State-Centred Data Construction and Training}
\label{sec:data}

We convert egocentric human video, third-person robot episodes and simulation into 3.7 million question--answer pairs. Apart from a small general-purpose replay, each pair poses one query of Table~\ref{tab:capabilities}.

Four rules govern the construction. \emph{State-dense queries}: queries are posed at multiple moments of one execution, covering sub-goals that are complete, in progress and not yet started. \emph{Annotation-derived answers}: a question is emitted only when its answer can be read from an annotation of the recording, which is stored with the pair. \emph{Same-task candidates}: anticipation distractors are steps of the current task, so only task order and the current observations separate completed steps from those ahead. \emph{Negative states}: successful demonstrations contain no failed outcomes, so failure recordings and simulation with known object states supply them.

\par\medskip
\noindent\textbf{Sources.}\quad
A sub-task annotation provides an observed structure $G$, and step time stamps determine how the state $S_t$ evolves. Egocentric human video contributes mostly anticipation and monitoring queries; third-person robot episodes, including failure recordings, cover all five capabilities; simulation contributes long task sequences. Appendix~\ref{app:composition} lists the datasets, question types and answer forms.

\begin{figure*}[!t]
\centering
\includegraphics[width=0.92\textwidth]{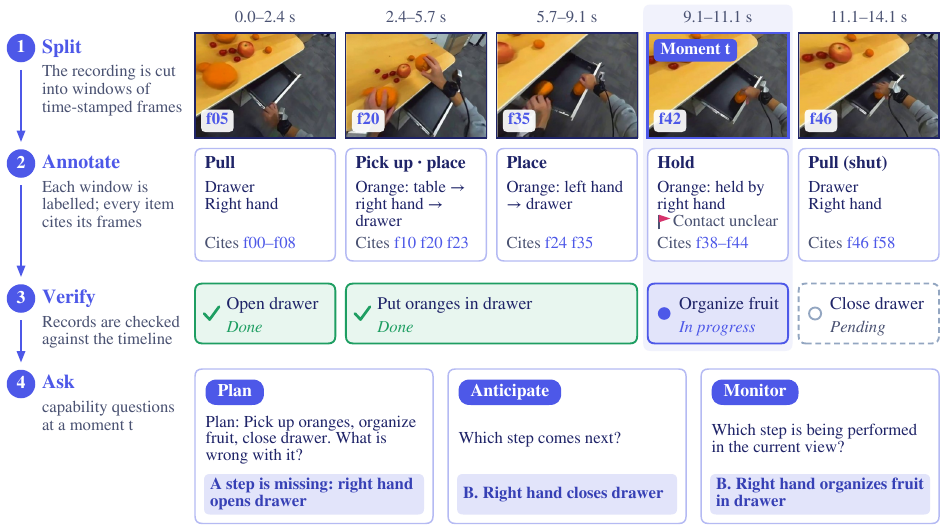}
\caption{\textbf{From egocentric video to annotation-derived questions}, on one head-camera recording of a person putting oranges into a drawer (conventions as in Figure~\ref{fig:overview}; each column is one window).
(1) The recording is cut into windows of time-stamped frames.
(2) A vision--language annotator writes a record per window, citing the frames that show every item.
(3) Records are compared with the recording's step timeline.
(4) At a sampled moment $t$ (shaded column), questions are derived from the timeline.}
\label{fig:pipeline}
\end{figure*}

\par\medskip
\noindent\textbf{Egocentric annotation.}\quad
Head-camera recordings rarely carry step boundaries or object states, so we annotate recordings beyond the public datasets ourselves (Figure~\ref{fig:pipeline}). For each window of a few seconds, a vision--language annotator receives time-stamped frames and writes a structured record of manipulation phases and object states. Every item cites its supporting frames, which gives the evidence $E_t^k$ of Eq.~\ref{eq:state}, and the annotator declines a window with insufficient evidence. Records are compared with the recording's step timeline, written independently of the annotator, and only content on which the two agree enters the dataset. Frames are sampled uniformly in time, not at step boundaries, so frame position does not reveal the answer.

\par\medskip
\noindent\textbf{Quality control and composition.}\quad
We control text shortcuts, answer-distribution biases and duplication per \emph{data family}, the pairs produced by one conversion from one source: a family answerable well above chance from question text alone is repaired or down-weighted, yes/no answers and answer letters are balanced, pairs with conflicting labels are removed, and the training data are deduplicated against the evaluation sets. Monitoring, anticipation and diagnosis account for about three quarters of the pairs (Figure~\ref{fig:composition}, Appendix~\ref{app:composition}), because one annotated recording yields a single plan but supports state queries at many moments. Third-person robot episodes supply 59\% of the pairs and egocentric video 32\%.

\par\medskip
\noindent\textbf{Training.}\quad
\ours is obtained from \base~\cite{qwen3.5} by supervised fine-tuning with the vision encoder frozen. The training data comprise three parts: 3.8 million task pairs (the collection and 0.14 million further pairs), 3.5 million spatial-grounding samples and 0.2 million progress-estimation samples. Training runs for one epoch of 11{,}835 optimisation steps, each of 64 packed sequences of 16{,}384 tokens, with the learning rate warmed up to $1.4\times10^{-5}$ over the first 355 steps and decayed to $1.4\times10^{-6}$ on a cosine schedule. The \emph{task-only model} used for the input controls of Section~\ref{sec:exp:grounding} is fine-tuned on the collection alone, for 6{,}582 steps with a warm-up of 184.

\section{Experiments}
\label{sec:exp}

\subsection{Setup}
\label{sec:exp:setup}

\par\medskip
\noindent\textbf{Benchmarks.}\quad
We map five public benchmarks onto the five queries as seventeen sub-items, by the principal judgment each requires (Table~\ref{tab:map}): \emph{EgoPlan-Bench}~\cite{egoplanbench}, four-way questions on the next action in egocentric video; the embodied-reasoning multiple-choice set of \emph{Cosmos-Reason1}~\cite{nvidia2025cosmos}; the planning dimension of \emph{RoboBench}~\cite{luo2025robobench}, with task decomposition (Q1), next sub-task (Q2) and step completion (Q3); \emph{LADDER}~\cite{petkov2026ladder}, with decomposition, verification, state and next sub-goal; and \emph{RoboFAC}~\cite{ye2025robofac}, with failure detection, identification and locating, and task identification. Each is scored from 0 to 100 by its native script, with a language-model judge for RoboBench Q1 and Q2 and RoboFAC task identification (Appendix~\ref{app:protocol}). RoboBench originally requires plans written as calls to a predefined skill library; we pose its Q1 and Q2 in natural language for every model, so our scores are not comparable with those reported for the original benchmark (Appendix~\ref{app:protocol}). We add \emph{ERIQ}~\cite{liu2025eriq}, the 3{,}905 questions of its eleven task-understanding subsets, scored by accuracy and not divided into sub-items; \emph{Avg6} is the unweighted mean of the six benchmarks. Progress estimation is measured on a 200-item benchmark with one item per episode: the model receives the task and four frames evenly spaced from the start of the episode to a target frame, and returns the overall task progress at that frame as a value in $[0,1]$. The reference progress gives each annotated stage of the episode a share of the task fixed by its type and advances linearly in time within a stage; MAE is the mean absolute error between prediction and reference.

\par\medskip
\noindent\textbf{Models.}\quad
We compare \ours with \astra and \opus; Qwen3.5 at 9B (our base) and 27B parameters; and twelve open embodied models of up to 32B parameters.

\subsection{Main Results}
\label{sec:exp:main}

\begin{table*}[!htb]
\centering
\caption{The six benchmarks (0--100), their unweighted mean (Avg6) and the progress-estimation error (MAE, lower is better). Best in \textbf{bold}, second best \underline{underlined}. }
\label{tab:main}
\small
\setlength{\tabcolsep}{2.8pt}
\resizebox{\textwidth}{!}{%
\begin{tabular}{lcccccccc}
\toprule
Model & EgoPlan-Bench & Cosmos-Reason1 & RoboBench & LADDER & RoboFAC & ERIQ & Avg6 & MAE\,$\downarrow$ \\
\midrule
\textbf{\ours (ours)} & \underline{74.5} & \textbf{83.5} & \underline{67.4} & \underline{79.0} & \textbf{69.2} & 85.1 & \textbf{76.5} & \textbf{0.087} \\
\midrule
\multicolumn{9}{l}{\emph{Closed models}} \\
GPT-6 Astra~\cite{openai2026gpt6astra} & \textbf{75.9} & \underline{77.5} & 66.0 & 75.7 & 53.2 & \underline{87.9} & 72.7 & 0.125 \\
Claude Opus 5.5~\cite{anthropic2026opus55} & 70.1 & 76.5 & \textbf{68.9} & \textbf{79.5} & \underline{57.7} & \textbf{88.4} & \underline{73.5} & \underline{0.118} \\
\midrule
\multicolumn{9}{l}{\emph{General-purpose open models}} \\
Qwen3.5-9B~\cite{qwen3.5} (base) & 48.8 & 62.0 & 55.6 & 68.8 & 35.9 & 71.4 & 57.1 & 0.257 \\
Qwen3.5-27B~\cite{qwen3.5} & 56.1 & 66.7 & 60.2 & 74.3 & 43.7 & 77.6 & 63.1 & 0.215 \\
\midrule
\multicolumn{9}{l}{\emph{Open embodied models}} \\
RoboBrain 2.5-8B~\cite{robobrain2_5} & 44.6 & 57.6 & 55.2 & 64.2 & 42.7 & 73.0 & 56.2 & 0.190 \\
RynnBrain 1.1-9B~\cite{dang2026} & 56.1 & 63.7 & 40.0 & 62.3 & 40.9 & 74.3 & 56.2 & 0.324 \\
SenseNova-SI 1.1-8B~\cite{cai2026sensenovasi} & 40.1 & 52.2 & 52.9 & 68.9 & 44.8 & 67.4 & 54.4 & 0.216 \\
PhysBrain 1.5-8B~\cite{physbrain15} & 66.9 & 72.7 & 51.4 & 48.5 & 38.6 & 72.1 & 58.4 & 0.218 \\
Embodied-R1.5-8B~\cite{yuan2026} & 60.5 & 68.6 & 50.4 & 59.6 & 43.6 & 67.6 & 58.4 & 0.280 \\
ACE-Brain-0.5-8B~\cite{acebrain05} & 31.1 & 47.3 & 45.7 & 45.4 & 41.5 & 61.4 & 45.4 & 0.333 \\
Cosmos-Reason1-7B~\cite{nvidia2025cosmos} & 33.3 & 61.7 & 52.4 & 50.9 & 38.1 & 63.9 & 50.1 & 0.274 \\
MiMo-Embodied-7B~\cite{mimo-embodied} & 42.1 & 60.3 & 54.7 & 70.2 & 34.7 & 67.2 & 54.9 & 0.239 \\
Hy-Embodied-VLM-1.0~\cite{hyembodiedvlm10} & 41.1 & 59.5 & 54.6 & 64.2 & 55.3 & 68.4 & 57.2 & 0.190 \\
RynnBrain-30B-A3B~\cite{dang2026} & 57.1 & 64.6 & 31.6 & 63.0 & 45.5 & 68.7 & 55.1 & 0.327 \\
RoboBrain 2.0-32B~\cite{robobrain2} & 51.4 & 53.5 & 56.1 & 65.0 & 39.5 & 67.8 & 55.6 & 0.267 \\
Cosmos-Reason2-32B~\cite{nvidia2025cosmosreason2} & 44.3 & 73.9 & 55.0 & 60.0 & 48.2 & 71.9 & 58.9 & 0.220 \\
\bottomrule
\end{tabular}}
\end{table*}

\ours has the highest Avg6 of all models: 76.5, against 72.7 for \astra and 73.5 for \opus (Table~\ref{tab:main}). It leads on Cosmos-Reason1 and RoboFAC, is second on EgoPlan-Bench, RoboBench and LADDER, within 1.5 points of the leading closed model on each, and is third on ERIQ; its progress MAE is 0.087, against 0.125 and 0.118. Qwen3.5-27B, three times the size of the base, is 6.0 points above it on Avg6; supervision at fixed size adds 19.4. Every open embodied model, including those of 30B to 32B parameters, averages below Qwen3.5-27B, and \ours is ahead of every open model on every benchmark.

\subsection{Capability Profile}
\label{sec:exp:capability}

\begin{figure*}[!t]
\centering
\includegraphics[width=0.9\textwidth]{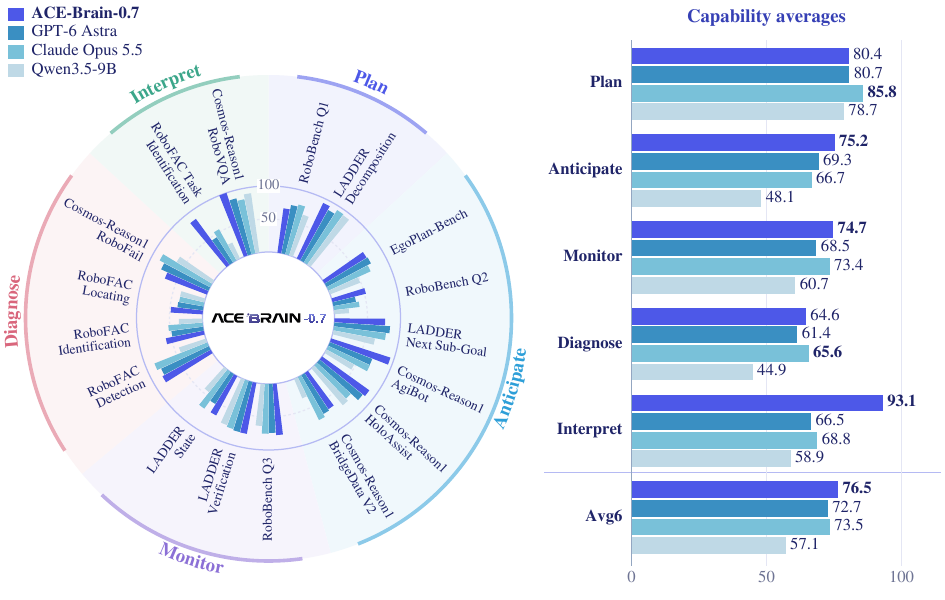}
\caption{\textbf{Capability profile.} \emph{Left:} scores of \ours, \astra, \opus and the base model on the seventeen sub-items, grouped by capability (values in Table~\ref{tab:subitems}); bar length is the score, with rings at 50 and 100. \emph{Right:} the unweighted mean over the sub-items of each capability, and Avg6 (Table~\ref{tab:main}).}
\label{fig:profile}
\end{figure*}

\begin{table*}[!htb]
\centering
\caption{\textbf{Structure and state in the same task families.} RoboBench by question type for the seven task families with at least 25 questions of each type. Best of the three models in \textbf{bold}.}
\label{tab:rb_family}
\footnotesize
\setlength{\tabcolsep}{3.6pt}
\begin{tabular}{l rccc rccc rccc}
\toprule
 & \multicolumn{4}{c}{Q1: task decomposition} & \multicolumn{4}{c}{Q2: next sub-task} & \multicolumn{4}{c}{Q3: step completion} \\
\cmidrule(lr){2-5}\cmidrule(lr){6-9}\cmidrule(lr){10-13}
Task family & $n$ & Ours & Astra & Opus & $n$ & Ours & Astra & Opus & $n$ & Ours & Astra & Opus \\
\midrule
Single arm & 235 & 85.7 & 84.5 & \textbf{88.3} & 231 & \textbf{61.0} & 35.5 & 38.2 & 367 & \textbf{78.7} & 76.6 & 78.2 \\
Rigid objects & 76 & 84.9 & 92.5 & \textbf{93.6} & 92 & \textbf{63.6} & 43.8 & 39.7 & 134 & 65.7 & \textbf{70.1} & 67.9 \\
Deformable objects & 114 & 69.0 & 78.6 & \textbf{84.3} & 103 & \textbf{51.3} & 34.0 & 44.7 & 137 & 78.1 & 73.7 & \textbf{79.6} \\
Articulated objects & 152 & 75.2 & 86.2 & \textbf{88.5} & 60 & 44.4 & 46.7 & \textbf{49.4} & 125 & \textbf{76.0} & 65.6 & 65.6 \\
Special materials & 119 & 71.8 & 82.4 & \textbf{87.3} & 90 & \textbf{53.5} & 32.0 & 38.9 & 119 & 73.9 & 73.1 & \textbf{74.8} \\
Multiple objects & 80 & 70.4 & 79.9 & \textbf{84.4} & 73 & \textbf{41.8} & 27.4 & 32.6 & 85 & \textbf{87.1} & 75.3 & 74.1 \\
World knowledge & 41 & 67.0 & 72.3 & \textbf{80.6} & 32 & \textbf{41.7} & 20.8 & 32.3 & 41 & \textbf{80.5} & 68.3 & 75.6 \\
\bottomrule
\end{tabular}
\end{table*}

\par\medskip
\noindent\textbf{Structure knowledge does not imply state knowledge.}\quad
On RoboBench, \astra and \opus lead \ours on task decomposition (74.5 and 77.8 against 67.7) and trail it on the next sub-task of the same episodes (34.7 and 39.0 against 52.1; Figure~\ref{fig:profile}): \opus is ahead by 10.1 points on the first and behind by 13.1 on the second. The reversal holds within task families (Table~\ref{tab:rb_family}): a closed model writes the best plan in all seven, and \ours names the next sub-task best in six. The planning gap to \opus grows with the length of the reference plan, from four points on plans of up to three steps to fifteen on plans of twelve or more (Figure~\ref{fig:errors}b; residual errors in Appendix~\ref{app:breakdowns}). \ours also leads on the next sub-task of the AgiBot source of Cosmos-Reason1 (94.3 against 69.0 and 69.5); on step completion the three models are within four points. On RoboFAC, \astra and \opus detect failures at 79.8 and 86.0 but locate them at 38.9 and 37.6, level with the 38.0 of \base; \ours locates them at 48.5. Failure locating and the next sub-task of a robot episode are the two lowest-scoring sub-items of all three models.

\par\medskip
\noindent\textbf{The gains are on queries that read the state.}\quad
Averaged over the sub-items of a capability, \ours gains 2 points over its base on planning, 14 on monitoring, 20 on diagnosis, 27 on anticipation and 34 on interpretation (Figure~\ref{fig:profile}, right). Planning is not saturated: on RoboBench Q1 the base is 13 points below \opus and \ours is 10 below.

\subsection{Task Prior versus Execution Grounding}
\label{sec:exp:grounding}

\begin{table}[!htb]
\centering
\caption{\textbf{Input controls.} Change in score (points) relative to the full input for the task-only model (Section~\ref{sec:data}) and \base (Task-only / Base). On RoboBench and LADDER the frame controls use the items with more than one image (2{,}024 and 1{,}028); Cosmos-Reason1 clips are run blind only.}
\label{tab:controls}
\footnotesize
\setlength{\tabcolsep}{5pt}
\begin{tabular}{lccccc}
\toprule
Input & EgoPlan-Bench & Cosmos-Reason1 & RoboBench & LADDER & RoboFAC \\
\midrule
Blind (no images) & $-47.8$ / $-21.1$ & $-41.2$ / $-24.1$ & $-11.1$ / $-2.7$ & $-17.5$ / $-14.2$ & $-25.0$ / $-6.5$ \\
First frame only & $-48.2$ / $-20.6$ & -- & $-20.9$ / $-6.5$ & $-9.9$ / $-14.2$ & $-0.5$ / $+7.3$ \\
Last frame only & $-12.1$ / $-1.2$ & -- & $-8.3$ / $-5.8$ & $-1.6$ / $+2.4$ & $-3.4$ / $+6.4$ \\
Shuffled frames & $-31.5$ / $-9.9$ & -- & $-17.2$ / $-5.0$ & $-1.8$ / $-3.8$ & $-6.5$ / $-1.7$ \\
\bottomrule
\end{tabular}
\end{table}

We test the two predictions of Section~\ref{sec:capabilities} with input controls (Table~\ref{tab:controls}; Appendix~\ref{app:protocol}) on \base and on the task-only model of Section~\ref{sec:data}, whose gains over the base on planning, monitoring and anticipation are 1.5, 13.9 and 26.1 points, against 1.8, 14.0 and 27.0 for \ours.

\par\medskip
\noindent\textbf{Structure is answered from the prior; state is not.}\quad
Without images, the task-only model loses 1.5 points on RoboBench task decomposition and 3.5 on LADDER decomposition, but 13.0 on the next sub-task and 25.3 on step completion. RoboFAC is the exception: the first frame alone scores within a point of all frames, so its questions are decided mainly by the scene.

\par\medskip
\noindent\textbf{Before training, next-step answers use the current view but not the history.}\quad
On EgoPlan-Bench both models fall to within three points of chance without images. \base loses 1.2 points with only the last frame and 9.9 with shuffled frames; the task-only model loses 12.1 and 31.5. A separate run of RoboBench Q2 that also varies whether the plan is given agrees (Appendix~\ref{app:protocol}). With no plan given, \base scores 22.4 with the frames and 26.2 without, and the task-only model 47.4 and 28.2. With the reference plan given as a numbered list and the answer an index into it, the task-only model selects the correct sub-task on 51.8\% of 674 questions and \base on 18.4\%, below the 33.4\% of the most frequent index.

\par\medskip
\noindent\textbf{Temporal question pairs.}\quad
We build 114 pairs from EgoPlan-Bench recordings: one question is posed at two moments of a recording, each 1.0\,s before the start of the step that is then correct, so only the state differs and a model that ignores the images cannot answer both. \base answers 47.4\% and 42.1\% of the two single questions correctly but both questions of a pair in 15.8\% of pairs. The task-only model answers 68.4\% and 71.1\% of the single questions and both in 54.4\% of pairs.

\begin{figure*}[!t]
\centering
\includegraphics[width=0.85\textwidth]{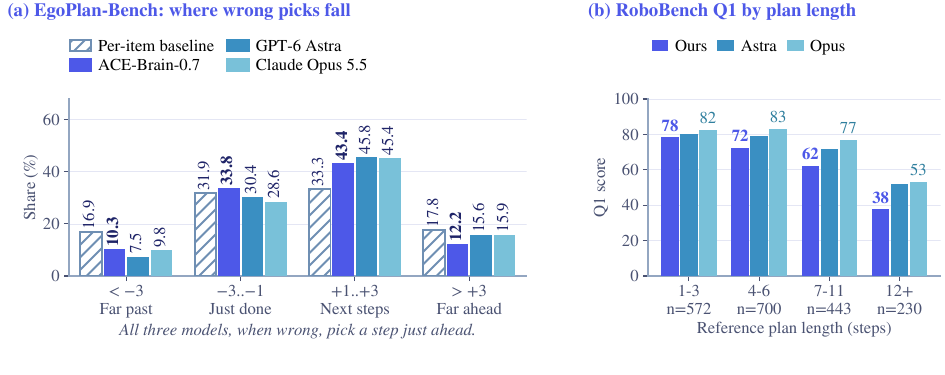}
\caption{\textbf{Error structure.} (a) EgoPlan-Bench (EPIC-KITCHENS): position of each model's wrong pick relative to the correct step on the timeline of the recording, for \ours, \astra and \opus; hatched bars give a uniform pick among each question's wrong options, averaged over the three models. (b) RoboBench Q1 score by reference-plan length.}
\label{fig:errors}
\end{figure*}

\par\medskip
\noindent\textbf{Wrong next steps lie just ahead of the true progress.}\quad
On EgoPlan-Bench, 99.5\% of the wrong options of the EPIC-KITCHENS questions are other steps of the same recording, so the structure is given. When \ours is wrong, 43.4\% of its picks are among the three steps after the correct one, 10.8 percentage points above a uniform pick among the distractors of the questions it answers wrongly; \astra and \opus show the same excess, 12.2 and 11.6 (Figure~\ref{fig:errors}a). On the three steps before the correct one, the excess of \ours is 0.5. \base shows the opposite pattern, with an excess of 1.2 on the three steps after the correct one and 7.4 on the three before: 50.8\% of its wrong answers repeat its answer from the first frame alone, against 28.0\% for the task-only model and 25\% by chance. On EgoPlan-Bench, the errors of the base repeat its first-frame answer; those of the three strongest models run ahead of the execution.

\subsection{Effect of Egocentric Data}
\label{sec:exp:ego}

\begin{figure}[!htb]
\centering
\includegraphics[width=0.8\textwidth]{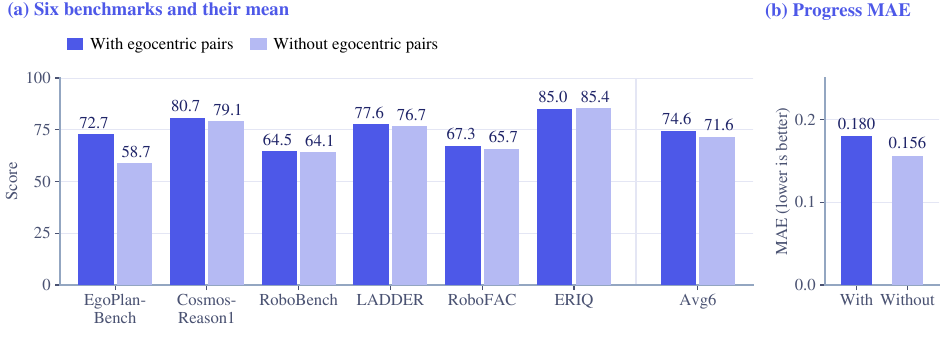}
\caption{\textbf{Effect of egocentric data.} Two models fine-tuned from \base on a quarter of the collection, with and without its egocentric pairs. (a) The six benchmarks and Avg6. (b) Progress MAE.}
\label{fig:ego_abl}
\end{figure}

Egocentric video supplies 32\% of the collection (Section~\ref{sec:data}). To measure its contribution, we fine-tune \base twice on a fixed quarter of the collection with the optimisation settings of Section~\ref{sec:data}: once on all 0.92 million pairs (1{,}676 steps) and once on the 0.62 million that remain when the egocentric pairs are removed (885 steps). Each model is trained for one epoch. The egocentric pairs raise Avg6 from 71.6 to 74.6 (Figure~\ref{fig:ego_abl}), and the gain is concentrated on EgoPlan-Bench, whose questions are posed on egocentric video: 72.7 against 58.7, a difference of 14.0 points. On the other five benchmarks the two models are within 1.6 points. The progress MAE is 0.156 without the egocentric pairs and 0.180 with them.

\subsection{Spatial Grounding}
\label{sec:exp:spatial}

Completion conditions are often spatial (the bowl is in the microwave), so reading $S_t$ involves locating objects. \ours is trained with spatial-grounding samples (Section~\ref{sec:data}); on the seven public spatial-grounding benchmarks of ACE-Brain-0.6~\cite{wang2026acebrain06} (Where2Place~\cite{yuan2024robopoint}, PixMo-Count~\cite{deitke2024molmopixmoopenweights}, RoboSpatial~\cite{robospatial}, RefSpatial~\cite{zhou2025roborefer}, PointArena~\cite{pointarena}, RoboAfford~\cite{roboafford} and ShareRobot~\cite{robobrain}), under its protocol, its Avg7 is 75.90, against 75.41 for ACE-Brain-0.6 and 73.68 for \astra, and it scores 91.25 on RefCOCOg~\cite{mao2016refcocog} (Table~\ref{tab:spatial}).

\begin{table*}[!htb]
\centering
\caption{\textbf{Spatial grounding.} The seven public benchmarks follow the protocol of~\cite{wang2026acebrain06}; Avg7 is their unweighted mean. RefCOCOg is Acc@0.5 and is not part of Avg7. Best in \textbf{bold}, second best \underline{underlined}.}
\label{tab:spatial}
\footnotesize
\setlength{\tabcolsep}{3.5pt}
\resizebox{\textwidth}{!}{%
\begin{tabular}{lccccccccc}
\toprule
Model & Where2Place & PixMo-Count & RoboSpatial & RefSpatial & PointArena & RoboAfford & ShareRobot & Avg7 & RefCOCOg \\
\midrule
\textbf{\ours (ours)} & \textbf{82.89} & \underline{89.98} & 73.13 & \underline{76.17} & \textbf{76.17} & \textbf{88.66} & 44.31 & \textbf{75.90} & \textbf{91.25} \\
\midrule
\multicolumn{10}{l}{\emph{Closed models}} \\
Claude Opus 5.5~\cite{anthropic2026opus55} & 66.44 & 83.74 & 73.30 & 71.10 & 61.20 & 75.10 & 40.20 & 67.30 & 85.89 \\
Gemini 4 Argon~\cite{googledeepmind2026gemini4argon} & 71.20 & 87.71 & \textbf{80.94} & 51.56 & 66.40 & 71.48 & 29.53 & 65.55 & -- \\
GPT-5.5~\cite{openai2026gpt55} & 45.25 & 82.61 & 59.19 & 60.29 & 68.02 & 68.04 & 38.57 & 60.28 & -- \\
GPT-5.6-sol~\cite{openai2026gpt56sol} & 46.23 & 83.55 & 64.34 & 54.51 & 69.76 & 71.45 & 36.66 & 60.93 & -- \\
GPT-6 Astra~\cite{openai2026gpt6astra} & 71.10 & 88.10 & 69.55 & \textbf{80.14} & \underline{75.36} & 84.94 & 46.60 & 73.68 & 89.05 \\
\midrule
\multicolumn{10}{l}{\emph{General-purpose open models}} \\
Qwen3-VL-8B~\cite{bai2025qwen3vltechnicalreport} & 62.60 & 86.40 & 62.66 & 54.15 & 61.81 & 74.83 & 35.80 & 62.61 & 88.94 \\
Qwen3.5-9B~\cite{qwen3.5} & 72.00 & \textbf{90.35} & 53.46 & 57.76 & 68.33 & 78.65 & 39.47 & 65.72 & \underline{89.98} \\
\midrule
\multicolumn{10}{l}{\emph{Open embodied models}} \\
RynnBrain-8B~\cite{dang2026} & 68.00 & 11.70 & 70.00 & 56.68 & 65.58 & 79.29 & 15.50 & 52.39 & 66.54 \\
RoboBrain 2.5-8B~\cite{robobrain2_5} & 64.00 & 78.80 & 63.71 & 59.21 & 59.27 & 76.33 & \textbf{56.24} & 65.37 & 69.53 \\
Embodied-R1.5-8B~\cite{yuan2026} & 77.00 & 81.70 & 68.96 & 57.04 & 71.38 & 83.43 & 37.21 & 68.10 & 86.30 \\
SenseNova-SI 1.1-8B~\cite{cai2026sensenovasi} & 47.00 & 59.20 & 59.65 & 29.78 & 51.83 & 50.71 & 25.98 & 46.31 & 81.84 \\
PhysBrain 1.5-8B~\cite{physbrain15} & 76.30 & 85.63 & 69.74 & 55.60 & 69.86 & 81.24 & 33.47 & 67.41 & 82.25 \\
ACE-Brain-0.6~\cite{wang2026acebrain06} & \underline{81.44} & 86.96 & \underline{76.56} & 71.84 & 74.13 & \underline{88.14} & \underline{48.82} & \underline{75.41} & 83.27 \\
\bottomrule
\end{tabular}}
\end{table*}

\section{Conclusion}
\label{sec:conclusion}

Describing a task by its structure and an execution by its state turns five public benchmarks into one profile of seventeen sub-items and separates what a model takes from the task prior from what it reads from the execution: plans survive the removal of the images; state answers do not. Supervised fine-tuning on 3.7 million annotation-derived question--answer pairs makes answers follow the execution shown. \ours, a 9B model obtained from \base by such fine-tuning on these pairs together with spatial-grounding and progress-estimation data, leads both closed models on the next sub-task of a robot episode, has the highest six-benchmark average of all compared models, and has the highest average of all compared models on seven public spatial-grounding benchmarks.

\clearpage
\bibliographystyle{unsrtnat}
\bibliography{references}

@article{pi07,
  author  = {{Physical Intelligence} and others},
  title   = {{$\pi_{0.7}$}: A Steerable Generalist Robotic Foundation Model with Emergent Capabilities},
  journal = {arXiv preprint arXiv:2604.15483},
  year    = {2026},
  url     = {https://arxiv.org/abs/2604.15483}
}

@misc{openai2026gpt6astra,
  author = {{OpenAI}},
  title  = {{GPT-6 Astra}: A New Generation of Intelligence},
  year   = {2026},
  month  = sep,
  url    = {https://openai.com/index/gpt-6-astra/}
}

@misc{qwen3.5,
  author = {{Qwen Team}},
  title  = {{Qwen3.5}: Towards Native Multimodal Agents},
  year   = {2026},
  month  = feb,
  url    = {https://qwen.ai/blog?id=qwen3.5}
}

@inproceedings{walke2023bridgedata,
  title     = {{BridgeData V2}: A Dataset for Robot Learning at Scale},
  author    = {Walke, Homer Rich and Black, Kevin and Lee, Abraham and Kim, Moo Jin and Du, Max and others},
  booktitle = {Conference on Robot Learning (CoRL)},
  year      = {2023},
  eprint = {2308.12952},
  archivePrefix = {arXiv}
}

@article{bai2025qwen3vltechnicalreport,
  title   = {{Qwen3-VL} Technical Report},
  author  = {Shuai Bai and Yuxuan Cai and Ruizhe Chen and Keqin Chen and Xionghui Chen and Zesen Cheng and Lianghao Deng and Wei Ding and Chang Gao and Chunjiang Ge and Wenbin Ge and Zhifang Guo and Qidong Huang and Jie Huang and Fei Huang and Binyuan Hui and Shutong Jiang and Zhaohai Li and Mingsheng Li and Mei Li and Kaixin Li and Zicheng Lin and Junyang Lin and Xuejing Liu and Jiawei Liu and Chenglong Liu and Yang Liu and Dayiheng Liu and Shixuan Liu and Dunjie Lu and Ruilin Luo and Chenxu Lv and Rui Men and Lingchen Meng and Xuancheng Ren and Xingzhang Ren and Sibo Song and Yuchong Sun and Jun Tang and Jianhong Tu and Jianqiang Wan and Peng Wang and Pengfei Wang and Qiuyue Wang and Yuxuan Wang and Tianbao Xie and Yiheng Xu and Haiyang Xu and Jin Xu and Zhibo Yang and Mingkun Yang and Jianxin Yang and An Yang and Bowen Yu and Fei Zhang and Hang Zhang and Xi Zhang and Bo Zheng and Humen Zhong and Jingren Zhou and Fan Zhou and Jing Zhou and Yuanzhi Zhu and Ke Zhu},
  year    = {2025},
  journal = {arXiv preprint arXiv:2511.21631}
}

@inproceedings{robobrain,
  title     = {{RoboBrain}: A Unified Brain Model for Robotic Manipulation from Abstract to Concrete},
  author    = {Ji, Yuheng and Tan, Huajie and Shi, Jiayu and Hao, Xiaoshuai and Zhang, Yuan and Zhang, Hengyuan and Wang, Pengwei and Zhao, Mengdi and Mu, Yao and An, Pengju and others},
  booktitle = {Proceedings of the Computer Vision and Pattern Recognition Conference (CVPR)},
  pages     = {1724--1734},
  year      = {2025},
  eprint = {2502.21257},
  archivePrefix = {arXiv}
}

@article{robobrain2,
  title   = {{RoboBrain 2.0} Technical Report},
  author  = {{BAAI RoboBrain Team} and Cao, Mingyu and Tan, Huajie and Ji, Yuheng and Chen, Xiansheng and Lin, Minglan and Li, Zhiyu and Cao, Zhou and Wang, Pengwei and Zhou, Enshen and others},
  journal = {arXiv preprint arXiv:2507.02029},
  year    = {2025}
}

@article{robobrain2_5,
  title   = {{RoboBrain 2.5}: Depth in Sight, Time in Mind},
  author  = {Tan, Huajie and Zhou, Enshen and Li, Zhiyu and Xu, Yijie and Ji, Yuheng and Chen, Xiansheng and Chi, Cheng and Wang, Pengwei and Jia, Huizhu and Ao, Yulong and others},
  journal = {arXiv preprint arXiv:2601.14352},
  year    = {2026}
}

@article{mimo-embodied,
  title   = {{MiMo-Embodied}: X-Embodied Foundation Model Technical Report},
  author  = {Hao, Xiaoshuai and Zhou, Lei and Huang, Zhijian and Hou, Zhiwen and Tang, Yingbo and Zhang, Lingfeng and Li, Guang and Lu, Zheng and Ren, Shuhuai and Meng, Xianhui and others},
  journal = {arXiv preprint arXiv:2511.16518},
  year    = {2025}
}

@article{cambrian,
  title   = {{Cambrian-1}: A Fully Open, Vision-Centric Exploration of Multimodal {LLMs}},
  author  = {Tong, Peter and Brown, Ellis and Wu, Penghao and Woo, Sanghyun and Iyer, Adithya Jairam Vedagiri and Akula, Sai Charitha and Yang, Shusheng and Yang, Jihan and Middepogu, Manoj and Wang, Ziteng and others},
  journal = {Advances in Neural Information Processing Systems},
  volume  = {37},
  pages   = {87310--87356},
  year    = {2024},
  eprint  = {2406.16860},
  archivePrefix = {arXiv}
}

@inproceedings{robovqa,
  title        = {{RoboVQA}: Multimodal Long-Horizon Reasoning for Robotics},
  author       = {Sermanet, Pierre and Ding, Tianli and Zhao, Jeffrey and Xia, Fei and Dwibedi, Debidatta and Gopalakrishnan, Keerthana and Chan, Christine and Dulac-Arnold, Gabriel and Maddineni, Sharath and Joshi, Nikhil J and others},
  booktitle    = {2024 IEEE International Conference on Robotics and Automation (ICRA)},
  pages        = {645--652},
  year         = {2024},
  organization = {IEEE},
  eprint = {2311.00899},
  archivePrefix = {arXiv}
}

@article{egoplanbench,
  title   = {{EgoPlan-Bench}: Benchmarking Multimodal Large Language Models for Human-Level Planning},
  author  = {Chen, Yi and Ge, Yuying and Ge, Yixiao and Ding, Mingyu and Li, Bohao and Wang, Rui and Xu, Ruifeng and Shan, Ying and Liu, Xihui},
  journal = {arXiv preprint arXiv:2312.06722},
  year    = {2023},
  eprint  = {2312.06722},
  archivePrefix = {arXiv}
}

@inproceedings{robospatial,
  title     = {{RoboSpatial}: Teaching Spatial Understanding to {2D} and {3D} Vision-Language Models for Robotics},
  author    = {Song, Chan Hee and Blukis, Valts and Tremblay, Jonathan and Tyree, Stephen and Su, Yu and Birchfield, Stan},
  booktitle = {Proceedings of the Computer Vision and Pattern Recognition Conference (CVPR)},
  pages     = {15768--15780},
  year      = {2025}
}

@inproceedings{roboafford,
  title     = {{RoboAfford}: A Dataset and Benchmark for Enhancing Object and Spatial Affordance Learning in Robot Manipulation},
  author    = {Tang, Yingbo and Zhang, Lingfeng and Zhang, Shuyi and Zhao, Yinuo and Hao, Xiaoshuai},
  booktitle = {Proceedings of the 33rd ACM International Conference on Multimedia},
  pages     = {12706--12713},
  year      = {2025}
}

@misc{nvidia2025cosmos,
  title   = {{Cosmos-Reason1}: From Physical Common Sense To Embodied Reasoning},
  author  = {{NVIDIA} and Azzolini, Alisson and others},
  howpublished = {arXiv preprint arXiv:2503.15558},
  year    = {2025},
  eprint  = {2503.15558},
  archivePrefix = {arXiv},
  url     = {https://arxiv.org/abs/2503.15558}
}

@inproceedings{yuan2024robopoint,
  title     = {{RoboPoint}: A Vision-Language Model for Spatial Affordance Prediction in Robotics},
  author    = {Yuan, Wentao and Duan, Jiafei and Blukis, Valts and Pumacay, Wilbert and Krishna, Ranjay and Murali, Adithyavairavan and Mousavian, Arsalan and Fox, Dieter},
  booktitle = {Proceedings of the 8th Conference on Robot Learning (CoRL)},
  series    = {Proceedings of Machine Learning Research},
  volume    = {270},
  pages     = {4005--4020},
  publisher = {PMLR},
  year      = {2025}
}

@inproceedings{mao2016refcocog,
  title     = {Generation and Comprehension of Unambiguous Object Descriptions},
  author    = {Mao, Junhua and Huang, Jonathan and Toshev, Alexander and Camburu, Oana and Yuille, Alan and Murphy, Kevin},
  booktitle = {Proceedings of the IEEE Conference on Computer Vision and Pattern Recognition (CVPR)},
  year      = {2016},
  eprint    = {1511.02283},
  archivePrefix = {arXiv}
}

@article{zhou2025roborefer,
  title   = {{RoboRefer}: Towards Spatial Referring with Reasoning in Vision-Language Models for Robotics},
  author  = {Zhou, Enshen and An, Jingkun and Chi, Cheng and Han, Yi and Rong, Shanyu and Zhang, Chi and Wang, Pengwei and Wang, Zhongyuan and Huang, Tiejun and Sheng, Lu and Zhang, Shanghang},
  journal = {Advances in Neural Information Processing Systems},
  volume  = {38},
  year    = {2025}
}

@article{lee2026roboreward,
  title   = {{RoboReward}: General-Purpose Vision-Language Reward Models for Robotics},
  author  = {Lee, Tony and Wagenmaker, Andrew and Pertsch, Karl and Liang, Percy and Levine, Sergey and Finn, Chelsea},
  journal = {arXiv preprint arXiv:2601.00675},
  year    = {2026},
  eprint  = {2601.00675},
  archivePrefix = {arXiv}
}

@article{hyembodiedvlm10,
  title   = {{Hy-Embodied-VLM-1.0}: Efficient Physical-World Agents},
  author  = {Wang, Ziyi and Yu, Xumin and Rao, Yongming and Ling, Yonggen and Li, Yunheng and Wang, Oran and Gao, Mingqi and Zhou, Yuchen and Liang, Yves and Liu, Zuyan and others},
  journal = {arXiv preprint arXiv:2607.12894},
  year    = {2026}
}

@misc{deitke2024molmopixmoopenweights,
  title         = {{Molmo} and {PixMo}: Open Weights and Open Data for State-of-the-Art Vision-Language Models},
  author        = {Matt Deitke and Christopher Clark and Sangho Lee and Rohun Tripathi and Yue Yang and Jae Sung Park and Mohammadreza Salehi and Niklas Muennighoff and Kyle Lo and Luca Soldaini and Jiasen Lu and Taira Anderson and Erin Bransom and Kiana Ehsani and Huong Ngo and YenSung Chen and Ajay Patel and Mark Yatskar and Chris Callison-Burch and Andrew Head and Rose Hendrix and Favyen Bastani and Eli VanderBilt and Nathan Lambert and Yvonne Chou and Arnavi Chheda and Jenna Sparks and Sam Skjonsberg and Michael Schmitz and Aaron Sarnat and Byron Bischoff and Pete Walsh and Chris Newell and Piper Wolters and Tanmay Gupta and Kuo-Hao Zeng and Jon Borchardt and Dirk Groeneveld and Crystal Nam and Sophie Lebrecht and Caitlin Wittlif and Carissa Schoenick and Oscar Michel and Ranjay Krishna and Luca Weihs and Noah A. Smith and Hannaneh Hajishirzi and Ross Girshick and Ali Farhadi and Aniruddha Kembhavi},
  year          = {2024},
  eprint        = {2409.17146},
  archiveprefix = {arXiv},
  primaryclass  = {cs.CV},
  url           = {https://arxiv.org/abs/2409.17146}
}

@misc{pointarena,
  title         = {{PointArena}: Probing Multimodal Grounding Through Language-Guided Pointing},
  author        = {Long Cheng and Jiafei Duan and Yi Ru Wang and Haoquan Fang and Boyang Li and Yushan Huang and Elvis Wang and Ainaz Eftekhar and Jason Lee and Wentao Yuan and Rose Hendrix and Noah A. Smith and Fei Xia and Dieter Fox and Ranjay Krishna},
  year          = {2025},
  eprint        = {2505.09990},
  archiveprefix = {arXiv},
  primaryclass  = {cs.CV},
  url           = {https://arxiv.org/abs/2505.09990}
}

@misc{dang2026,
  title = {{RynnBrain: Open Embodied Foundation Models}},
  author = {Ronghao Dang and Jiayan Guo and Bohan Hou and Sicong Leng and Kehan Li and Xin Li and Jiangpin Liu and Yunxuan Mao and Zhikai Wang and Yuqian Yuan and Minghao Zhu and Xiao Lin and Yang Bai and Qian Jiang and Yaxi Zhao and Minghua Zeng and Junlong Gao and Yuming Jiang and Jun Cen and Siteng Huang and Liuyi Wang and Wenqiao Zhang and Chengju Liu and Jianfei Yang and Shijian Lu and Deli Zhao},
  year = {2026},
  howpublished = {arXiv preprint arXiv:2602.14979},
  eprint = {2602.14979},
  archivePrefix = {arXiv},
  primaryClass = {cs.RO},
  url = {https://arxiv.org/abs/2602.14979}
}

@misc{yuan2026,
  title = {{Embodied-R1.5: Evolving Physical Intelligence via Embodied Foundation Models}},
  author = {Yifu Yuan and Yaoting Huang and Xianze Yao and Yutong Li and Shuoheng Zhang and Linqi Han and Pengyi Li and Jiangeng Sun and Wenting Jia and Zhao Zhang and Yuhao Liu and Ruihao Liao and Yucheng Hu and Qiyu Wu and Yuxiao Li and Zibin Dong and Fei Ni and Yan Zheng and Shuyang Gu and Yi Ma and Hongyao Tang and Han Hu and Jianye Hao},
  year = {2026},
  howpublished = {arXiv preprint arXiv:2606.11324},
  eprint = {2606.11324},
  archivePrefix = {arXiv},
  primaryClass = {cs.RO},
  url = {https://arxiv.org/abs/2606.11324}
}

@inproceedings{cai2026sensenovasi,
  title     = {Scaling Spatial Intelligence with Multimodal Foundation Models},
  author    = {Cai, Zhongang and Wang, Ruisi and Gu, Chenyang and Pu, Fanyi and Xu, Junxiang and Wang, Yubo and Yin, Wanqi and Yang, Zhitao and Wei, Chen and Zhou, Tongxi and Sun, Qingping and Pang, Hui En and Li, Jiaqi and Qian, Oscar and Lin, Zhiqian and Shi, Xuanke and Deng, Kewang and Han, Xiaoyang and Chen, Zukai and Fan, Xiangyu and Deng, Hanming and Lu, Lewei and Pan, Liang and Li, Bo and Liu, Ziwei and Wang, Quan and Lin, Dahua and Yang, Lei},
  booktitle = {Proceedings of the IEEE/CVF Conference on Computer Vision and Pattern Recognition (CVPR)},
  pages     = {7879--7890},
  year      = {2026},
  eprint    = {2511.13719},
  archivePrefix = {arXiv}
}

@article{physbrain15,
  title   = {{PhysBrain 1.5}: From Vision-Language Models to Physical Foundation Models},
  author  = {{DeepCybo Team} and Bin, Yu and Cao, Haipeng and Chang, Zheng and Chen, Kai and Chen, Youning and Deng, Kailin and Du, Yichao and Fu, Xiaotong and Ge, Haoyang and others},
  journal = {arXiv preprint arXiv:2609.14973},
  year    = {2026}
}

@misc{anthropic2026opus55,
  title        = {System Card: {Claude Opus 5.5}},
  author       = {{Anthropic}},
  month        = {September},
  year         = {2026},
  howpublished = {\url{https://www.anthropic.com/claude-opus-5-5-system-card}}
}

@misc{googledeepmind2026gemini4argon,
  author       = {{Google DeepMind}},
  title        = {{Gemini 4 Argon}: Model Card},
  year         = {2026},
  month        = oct,
  url          = {https://deepmind.google/models/model-cards/gemini-4-argon/},
  note         = {Published October 13, 2026}
}

@misc{openai2026gpt55,
  author = {{OpenAI}},
  title  = {{GPT-5.5 Model}},
  year   = {2026},
  url    = {https://developers.openai.com/api/docs/models/gpt-5.5},
  note   = {OpenAI API documentation}
}

@misc{openai2026gpt56sol,
  author = {{OpenAI}},
  title  = {{GPT-5.6 Sol Model}},
  year   = {2026},
  url    = {https://developers.openai.com/api/docs/models/gpt-5.6-sol},
  note   = {OpenAI API documentation}
}

@inproceedings{oxe2023,
  title   = {Open X-Embodiment: Robotic Learning Datasets and {RT-X} Models},
  author  = {{Open X-Embodiment Collaboration} and others},
  booktitle = {IEEE International Conference on Robotics and Automation (ICRA)},
  year    = {2024},
  url     = {https://arxiv.org/abs/2310.08864}
}

@article{liu2025eriq,
  title   = {Unified Embodied {VLM} Reasoning with Robotic Action via Autoregressive Discretized Pre-training},
  author  = {Liu, Yi and Wang, Sukai and Wei, Dafeng and Cai, Xiaowei and Zhong, Linqing and Yang, Jiange and Ren, Guanghui and Zhang, Jinyu and Yao, Maoqing and Li, Chuankang and He, Xindong and Chen, Liliang and Luo, Jianlan},
  journal = {arXiv preprint arXiv:2512.24125},
  year    = {2025},
  eprint  = {2512.24125},
  archivePrefix = {arXiv},
  url     = {https://arxiv.org/abs/2512.24125}
}

@article{wang2026acebrain06,
  title   = {{ACE-Brain-0.6}: Scaling Spatial Grounding for Physical Foundation Models},
  author  = {Wang, Yinhe and Tao, Tao and He, Zexin and Luo, Shuang and Mao, Wei and Mu, Yao and Pan, Feng and Deng, Hanming and Lu, Lewei and Hou, Zhi and Pan, Xingang and Lin, Dahua and Tao, Dacheng and Wang, Xiaogang},
  journal = {arXiv preprint arXiv:2610.27040},
  year    = {2026},
  eprint  = {2610.27040},
  archivePrefix = {arXiv},
  url     = {https://arxiv.org/abs/2610.27040}
}

@article{petkov2026ladder,
  title         = {{LADDER}: A Long-Horizon Robot Manipulation Benchmark},
  author        = {Petkov, Petko and Albanese, Giuliano and Nikolov, Nikolay and Van Gool, Luc and Tremblay, Jonathan and Paudel, Danda Pani},
  journal       = {arXiv preprint arXiv:2610.18053},
  year          = {2026},
  eprint        = {2610.18053},
  archivePrefix = {arXiv}
}

@article{luo2025robobench,
  title         = {{RoboBench}: A Comprehensive Evaluation Benchmark for Multimodal Large Language Models as Embodied Brain},
  author        = {Luo, Yulin and Fan, Chun-Kai and Dong, Menghang and Shi, Jiayu and Mi, Xiangju and Zhao, Mengdi and Zhang, Bo-Wen and Chi, Cheng and Liu, Jiaming and Dai, Gaole and Zhang, Rongyu and An, Ruichuan and Wu, Kun and Che, Zhengping and Xie, Shaoxuan and Yao, Guocai and Zhao, Zhongxia and Wang, Pengwei and Liu, Guang and Wang, Zhongyuan and Huang, Tiejun and Zhang, Shanghang},
  journal       = {arXiv preprint arXiv:2510.17801},
  year          = {2025},
  eprint        = {2510.17801},
  archivePrefix = {arXiv}
}

@article{ye2025robofac,
  title         = {{RoboFAC}: A Comprehensive Framework for Robotic Failure Analysis and Correction},
  author        = {Ye, Zewei and Lu, Weifeng and Ye, Minghao and Lin, Tao and Yang, Shuo and Yan, Junchi and Zhao, Bo},
  journal       = {arXiv preprint arXiv:2505.12224},
  year          = {2025},
  eprint        = {2505.12224},
  archivePrefix = {arXiv}
}

@misc{nvidia2025cosmosreason2,
  author = {{NVIDIA}},
  title  = {{Cosmos-Reason2}: Open Reasoning Vision-Language Model for Physical {AI}},
  year   = {2025},
  howpublished = {NVIDIA Cosmos documentation},
  url    = {https://docs.nvidia.com/cosmos/latest/reason2/index.html}
}

@article{bu2025agibot,
  title         = {{AgiBot World Colosseo}: A Large-scale Manipulation Platform for Scalable and Intelligent Embodied Systems},
  author        = {{AgiBot World Contributors} and Bu, Qingwen and Cai, Jisong and Chen, Li and Cui, Xiuqi and others},
  journal       = {arXiv preprint arXiv:2503.06669},
  year          = {2025},
  eprint        = {2503.06669},
  archivePrefix = {arXiv}
}

@inproceedings{khazatsky2024droid,
  title         = {{DROID}: A Large-Scale In-The-Wild Robot Manipulation Dataset},
  author        = {Khazatsky, Alexander and Pertsch, Karl and Nair, Suraj and Balakrishna, Ashwin and Dasari, Sudeep and others},
  booktitle       = {Robotics: Science and Systems (RSS)},
  year          = {2024},
  eprint        = {2403.12945},
  archivePrefix = {arXiv}
}

@article{damen2020epic,
  title         = {Rescaling Egocentric Vision: Collection, Pipeline and Challenges for {EPIC-KITCHENS-100}},
  author        = {Damen, Dima and Doughty, Hazel and Farinella, Giovanni Maria and Furnari, Antonino and Kazakos, Evangelos and others},
  journal       = {International Journal of Computer Vision},
  year          = {2022},
  volume        = {130},
  pages         = {33--55},
  eprint        = {2006.13256},
  archivePrefix = {arXiv}
}

@inproceedings{grauman2021ego4d,
  title         = {{Ego4D}: Around the World in 3,000 Hours of Egocentric Video},
  author        = {Grauman, Kristen and Westbury, Andrew and Byrne, Eugene and Chavis, Zachary and Furnari, Antonino and others},
  booktitle       = {Proceedings of the IEEE/CVF Conference on Computer Vision and Pattern Recognition (CVPR)},
  year          = {2022},
  eprint        = {2110.07058},
  archivePrefix = {arXiv}
}

@inproceedings{wang2023holoassist,
  title         = {{HoloAssist}: an Egocentric Human Interaction Dataset for Interactive {AI} Assistants in the Real World},
  author        = {Wang, Xin and Kwon, Taein and Rad, Mahdi and Pan, Bowen and Chakraborty, Ishani and others},
  booktitle       = {Proceedings of the IEEE/CVF International Conference on Computer Vision (ICCV)},
  year          = {2023},
  eprint        = {2309.17024},
  archivePrefix = {arXiv}
}

@article{nasiriany2024robocasa,
  title         = {{RoboCasa}: Large-Scale Simulation of Everyday Tasks for Generalist Robots},
  author        = {Nasiriany, Soroush and Maddukuri, Abhiram and Zhang, Lance and Parikh, Adeet and Lo, Aaron and others},
  journal       = {arXiv preprint arXiv:2406.02523},
  year          = {2024},
  eprint        = {2406.02523},
  archivePrefix = {arXiv}
}

@article{zeng2025vifailback,
  title         = {Diagnose, Correct, and Learn from Manipulation Failures via Visual Symbols},
  author        = {Zeng, Xianchao and Zhou, Xinyu and Li, Youcheng and Shi, Jiayou and Li, Tianle and others},
  journal       = {arXiv preprint arXiv:2512.02787},
  year          = {2025},
  eprint        = {2512.02787},
  archivePrefix = {arXiv},
  note          = {ViFailback; accepted to CVPR 2026}
}

@article{zhang2026progresslm,
  title         = {{ProgressLM}: Towards Progress Reasoning in Vision-Language Models},
  author        = {Zhang, Jianshu and Qian, Chengxuan and Sun, Haosen and Lu, Haoran and Wang, Dingcheng and Xue, Letian and Liu, Han},
  journal       = {arXiv preprint arXiv:2601.15224},
  year          = {2026},
  eprint        = {2601.15224},
  archivePrefix = {arXiv}
}

@inproceedings{ma2025gvl,
  title         = {Vision Language Models are In-Context Value Learners},
  author        = {Ma, Yecheng Jason and Hejna, Joey and Wahid, Ayzaan and Fu, Chuyuan and Shah, Dhruv and Liang, Jacky and Xu, Zhuo and Kirmani, Sean and Xu, Peng and Driess, Danny and Xiao, Ted and Tompson, Jonathan and Bastani, Osbert and Jayaraman, Dinesh and Yu, Wenhao and Zhang, Tingnan and Sadigh, Dorsa and Xia, Fei},
  booktitle       = {International Conference on Learning Representations (ICLR)},
  year          = {2025},
  eprint        = {2411.04549},
  archivePrefix = {arXiv}
}

@inproceedings{duan2025aha,
  title         = {{AHA}: A Vision-Language-Model for Detecting and Reasoning Over Failures in Robotic Manipulation},
  author        = {Duan, Jiafei and Pumacay, Wilbert and Kumar, Nishanth and Wang, Yi Ru and Tian, Shulin and Yuan, Wentao and Krishna, Ranjay and Fox, Dieter and Mandlekar, Ajay and Guo, Yijie},
  booktitle       = {International Conference on Learning Representations (ICLR)},
  year          = {2025},
  eprint        = {2410.00371},
  archivePrefix = {arXiv}
}

@article{pacaud2025guardian,
  title         = {Guardian: Detecting Robotic Manipulation Failures with Vision-Language Models},
  author        = {Pacaud, Paul and Garcia, Ricardo and Chen, Shizhe and Schmid, Cordelia},
  journal       = {arXiv preprint arXiv:2512.01946},
  year          = {2025},
  eprint        = {2512.01946},
  archivePrefix = {arXiv}
}

@article{shi2025hirobot,
  title         = {{Hi Robot}: Open-Ended Instruction Following with Hierarchical Vision-Language-Action Models},
  author        = {Shi, Lucy Xiaoyang and Ichter, Brian and Equi, Michael and Ke, Liyiming and Pertsch, Karl and Vuong, Quan and Tanner, James and Walling, Anna and Wang, Haohuan and Fusai, Niccolo and Li-Bell, Adrian and Driess, Danny and Groom, Lachy and Levine, Sergey and Finn, Chelsea},
  journal       = {arXiv preprint arXiv:2502.19417},
  year          = {2025},
  eprint        = {2502.19417},
  archivePrefix = {arXiv},
  note = {ICML 2025}
}

@misc{geminirobotics15_2025,
  title         = {{Gemini Robotics 1.5}: Pushing the Frontier of Generalist Robots with Advanced Embodied Reasoning, Thinking, and Motion Transfer},
  author        = {{Gemini Robotics Team}},
  year          = {2025},
  eprint        = {2510.03342},
  archivePrefix = {arXiv},
  howpublished  = {arXiv preprint arXiv:2510.03342}
}

@inproceedings{liu2023reflect,
  title     = {{REFLECT}: Summarizing Robot Experiences for Failure Explanation and Correction},
  author    = {Liu, Zeyi and Bahety, Arpit and Song, Shuran},
  booktitle = {Conference on Robot Learning (CoRL)},
  year      = {2023},
  eprint    = {2306.15724},
  archivePrefix = {arXiv}
}

@article{acebrain05,
  title   = {{ACE-Brain-0.5}: A Unified Embodied Foundational Model for Physical Agentic {AI}},
  author  = {{ACE-Brain Team} and Gong, Ziyang and Gu, Haoming and Luo, Zehang and Zhang, Tianyi and Tao, Tao and others},
  journal = {arXiv preprint arXiv:2607.04426},
  year    = {2026},
  eprint  = {2607.04426},
  archivePrefix = {arXiv}
}

\appendix
\section{Appendix}
\subsection{Contributions and Author List}
\label{app:contributors}

\noindent\textbf{Contributors:}
Yinhe Wang\textsuperscript{1,*},
Shijun Shi\textsuperscript{1,3,*},
Feng Pan\textsuperscript{1,*},
Tao Tao\textsuperscript{1},
Wei Mao\textsuperscript{1},
Jiangning Zhang\textsuperscript{3},
Zhiyao Ren\textsuperscript{1},
Yihang Jiao\textsuperscript{1},
Haoming Gu\textsuperscript{1},
Yunpeng Qing\textsuperscript{1},
Jiyu Guo\textsuperscript{1}.

\medskip
\noindent\textbf{Advisors:}
Zhi Hou\textsuperscript{1},
Dacheng Tao\textsuperscript{2},
Xiaogang Wang\textsuperscript{1}.

\medskip
\noindent\textbf{Affiliations:}\\
\textsuperscript{1}ACE Robotics\\
\textsuperscript{2}Nanyang Technological University\\
\textsuperscript{3}Zhejiang University

\medskip
{\footnotesize
\noindent
\textsuperscript{*} Equal contribution.\par
}

\subsection{Evaluation Protocol Notes}
\label{app:protocol}

\par\medskip
\noindent\textbf{Decoding and inputs.}\quad
All open models, including \ours and the base model, decode greedily; on Cosmos-Reason1 the generation budget is 4{,}096 tokens with a repetition penalty of 1.05. Cosmos-Reason1 and RoboBench images are undistorted for every model. RoboBench is used in a natural-language version for every model: in Q1 and Q2 the skill library and worked examples are removed from the prompt, the model answers with numbered steps in plain language instead of skill-function calls, and the reference plans are converted from function calls to sentences by fixed templates; Q3 is unchanged. Navigation questions and 28 questions whose converted references could not be verified are excluded, leaving 3{,}979 items, and the reported score is the mean over these items.

\par\medskip
\noindent\textbf{Scoring.}\quad
Multiple-choice and yes/no answers are extracted and scored by each benchmark's native script. On EgoPlan-Bench the open models are scored from the option letter at the end of the response and the closed models with the extractor of the public evaluation toolkit. A language-model judge (Qwen3.8-Flash) scores RoboBench Q1 and Q2 and RoboFAC task identification against the reference, with the same prompt for every model; RoboBench Q3 is scored as yes/no. LADDER is the mean of decomposition exact match, verification balanced accuracy, state accuracy and next-sub-goal validity; its state component counts a scene as correct only if every state query about it is correct. RoboFAC is the macro average of its four question types. The progress MAE counts an unparsable answer as an error of 1. Sub-item scores of the closed models are computed from their per-item predictions.

\par\medskip
\noindent\textbf{Input controls and temporal question pairs.}\quad
The controls of Section~\ref{sec:exp:grounding} change only the images of a question; items with a single image are excluded from the three frame controls. The plan-given comparison is a separate run of the 842 Q2 questions, so its scores differ from those of Tables~\ref{tab:controls} and~\ref{tab:subitems}; its plan is the reference Q1 plan of the same episode, aligned by index for 674 of them. A temporal question pair comes from an EgoPlan-Bench question on EPIC-KITCHENS whose options contain two consecutive steps of the recording; each cut point lies 1.0\,s before the start of one of them.

\subsection{Construction and Composition of the Training Collection}
\label{app:composition}

\begin{figure*}[!htb]
    \centering
    \includegraphics[width=0.85\textwidth]{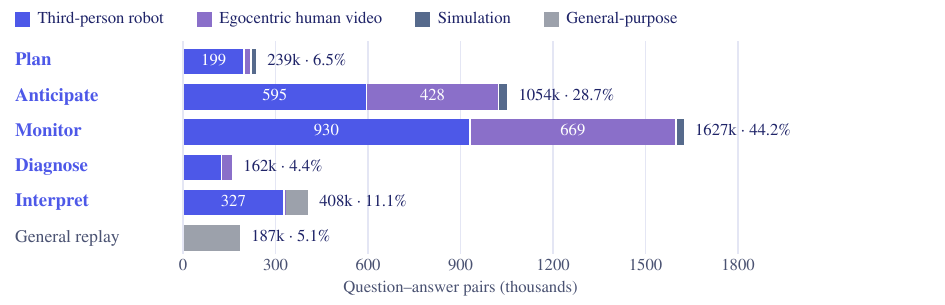}
    \caption{\textbf{Composition of the training collection}: pairs per capability, divided by source, with the count and share at the end of each bar (counts in Table~\ref{tab:comp_matrix}).}
    \label{fig:composition}
\end{figure*}

Table~\ref{tab:datasets} lists the datasets of each kind of recording; Table~\ref{tab:construction}, the annotations, question types and answer form of each capability; and Table~\ref{tab:comp_matrix}, the composition by capability and source plotted in Figure~\ref{fig:composition}. One shortcut found in quality control is a correct option that is systematically the longest.

\par\medskip
\noindent\textbf{Annotation record of the egocentric pipeline.}\quad
We keep the spans in which the wearer's hands act on objects and cut them into windows of a few seconds, sampled densely enough that no grasp or release falls between frames. The record an annotator writes for a window (Section~\ref{sec:data}) contains a status (annotated, or declined for no manipulation or insufficient evidence), the objects involved, the manipulations divided into phases (verb, acting hand, object roles, first and last frame), object states in time order, relations such as contact and containment, and declared uncertainties. Every item carries the identifiers of the frames that show it, and validation rejects a record with an identifier outside the window or with unordered phase boundaries. A state-change question must show the frames cited for both its before and its after observation.

\subsection{Additional Breakdowns and Cases}
\label{app:breakdowns}

Table~\ref{tab:map} maps benchmark sub-items to capabilities. Two sub-items mix query types: the RoboFail~\cite{liu2023reflect} questions of Cosmos-Reason1 cover success judgments and next-step feasibility, and RoboVQA~\cite{robovqa} spans affordance, success and planning. To keep the benchmarks' scoring units, we assign them to diagnosis and interpretation, respectively. Table~\ref{tab:subitems} gives the sub-item scores plotted in Figure~\ref{fig:profile}. Tables~\ref{tab:rb_q1} and~\ref{tab:egoplan_subsets} break down RoboBench Q1 by embodiment, object attribute, camera views and reference plan length, and EgoPlan-Bench by source, by whether the frames show progress, and by the length of the correct option. The task families of Table~\ref{tab:rb_family} are those with at least 25 planning questions of each type; they cover 681 of the 842 Q2 questions and 1{,}008 of the 1{,}192 Q3 questions.

\par
\begin{figure*}[!htb]
\centering
\includegraphics[width=0.9\textwidth]{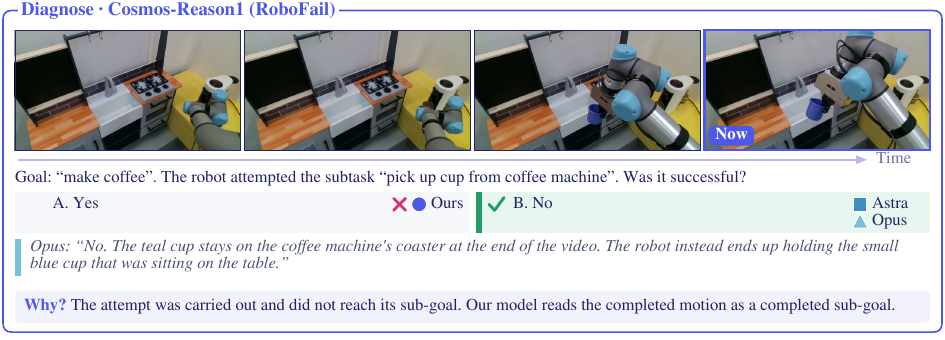}
\caption{\textbf{A negative state that \ours does not recognise} (Cosmos-Reason1, RoboFail). The robot was to make coffee and attempted to pick up the cup from the coffee machine; it ended up holding a different cup. \ours judges the sub-task successful and \opus answers correctly.}
\label{fig:case_limit}
\end{figure*}

\medskip
\noindent\textbf{Residual errors.}\quad
A closed model leads \ours by more than three points on four sub-items (Table~\ref{tab:subitems}); we examine three. On the LADDER next sub-goal, where \astra leads, the most frequent error of \ours is to issue a sub-goal that is already done: it scores 68.0 on the 337 items with at least one completed sub-task, against 75.4 for \astra and 70.6 for \opus. On the RoboFail source of Cosmos-Reason1, where \opus leads, \ours answers ``yes'' to 84.0\% of the questions although the reference is ``yes'' for 58\%; when the reference is ``no'', it is correct in 31.0\% of cases, against 69.0\% for \astra and 78.6\% for \opus (Figure~\ref{fig:case_limit}). On RoboBench task decomposition, where \opus leads, \ours writes 5.6 steps on average, the reference has 6.7 and \opus writes 7.4. Most long reference plans repeat a grasp-and-place cycle once per object: on the 117 questions about a counted set of objects \ours scores 36.8 against 58.7 for \opus, and on the 157 dual-arm questions 36.5 against 64.9. On the Q1 subsets whose instruction singles out an object by shape, colour or size, \ours is ahead of both closed models (Table~\ref{tab:rb_q1}).

\par
\begin{figure*}[!htb]
\centering
\includegraphics[width=0.9\textwidth]{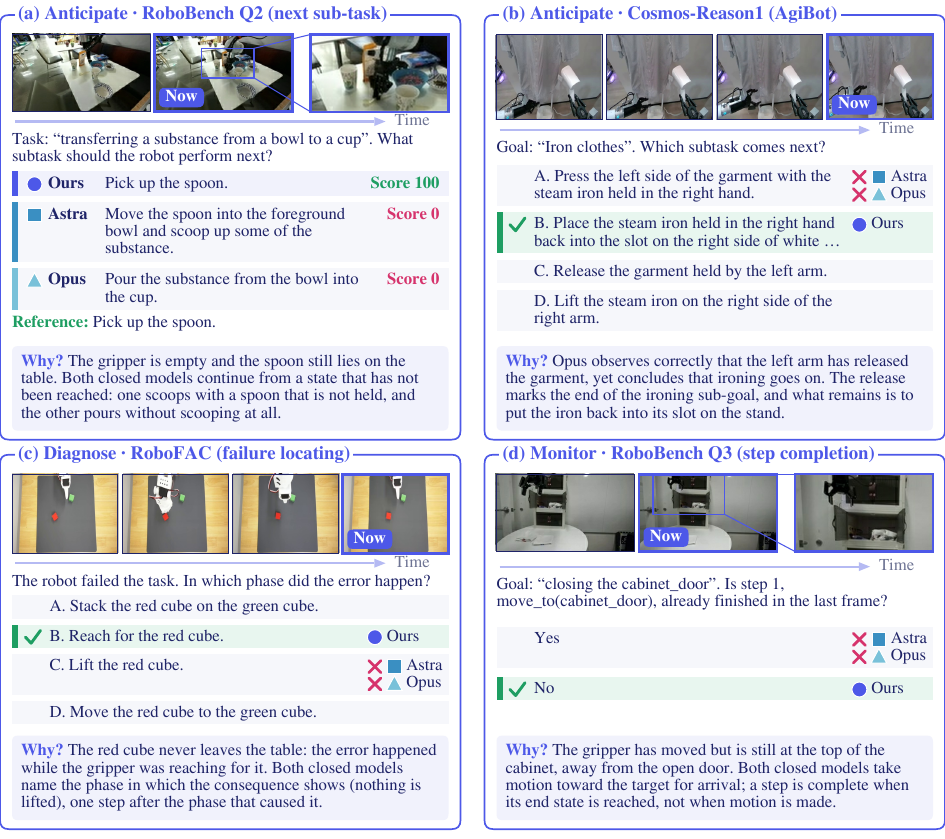}
\caption{\textbf{Reading state from robot recordings: four benchmark items} (``Ours'': the task-only model). Each card shows the frames given to the models (the last is the current view), the question and each model's answer; the reference is in green. (a) RoboBench Q2 and (b) Cosmos-Reason1 (AgiBot): the next sub-task. (c) RoboFAC: the phase in which the failure occurred. (d) RoboBench Q3: whether a step is complete.}
\label{fig:cases}
\end{figure*}

\begin{figure*}[!htb]
\centering
\includegraphics[width=0.85\textwidth]{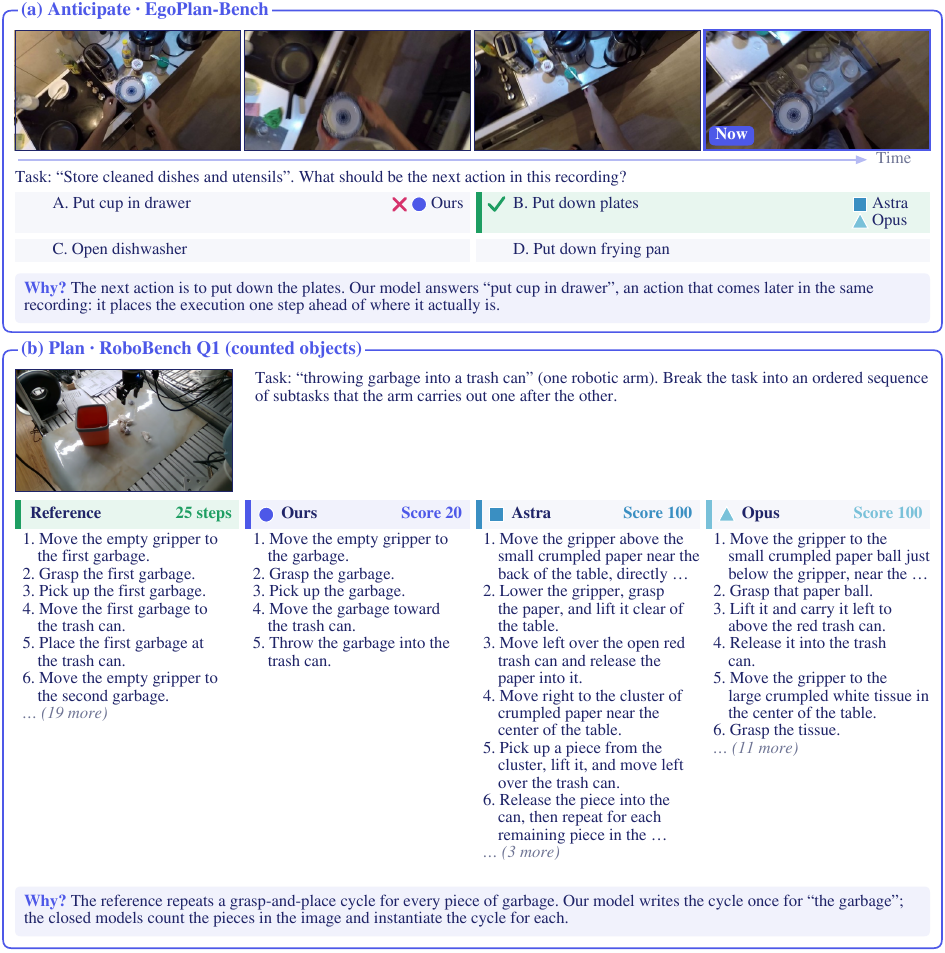}
\caption{\textbf{Two cases in which the task-only model is wrong.} (a) EgoPlan-Bench: it skips a step that has not yet been carried out. (b) RoboBench Q1: it writes one grasp-and-place cycle for a task that requires one per object.}
\label{fig:case_limits_more}
\end{figure*}

\begin{figure*}[!htb]
\centering
\includegraphics[width=0.9\textwidth,height=0.8\textheight,keepaspectratio]{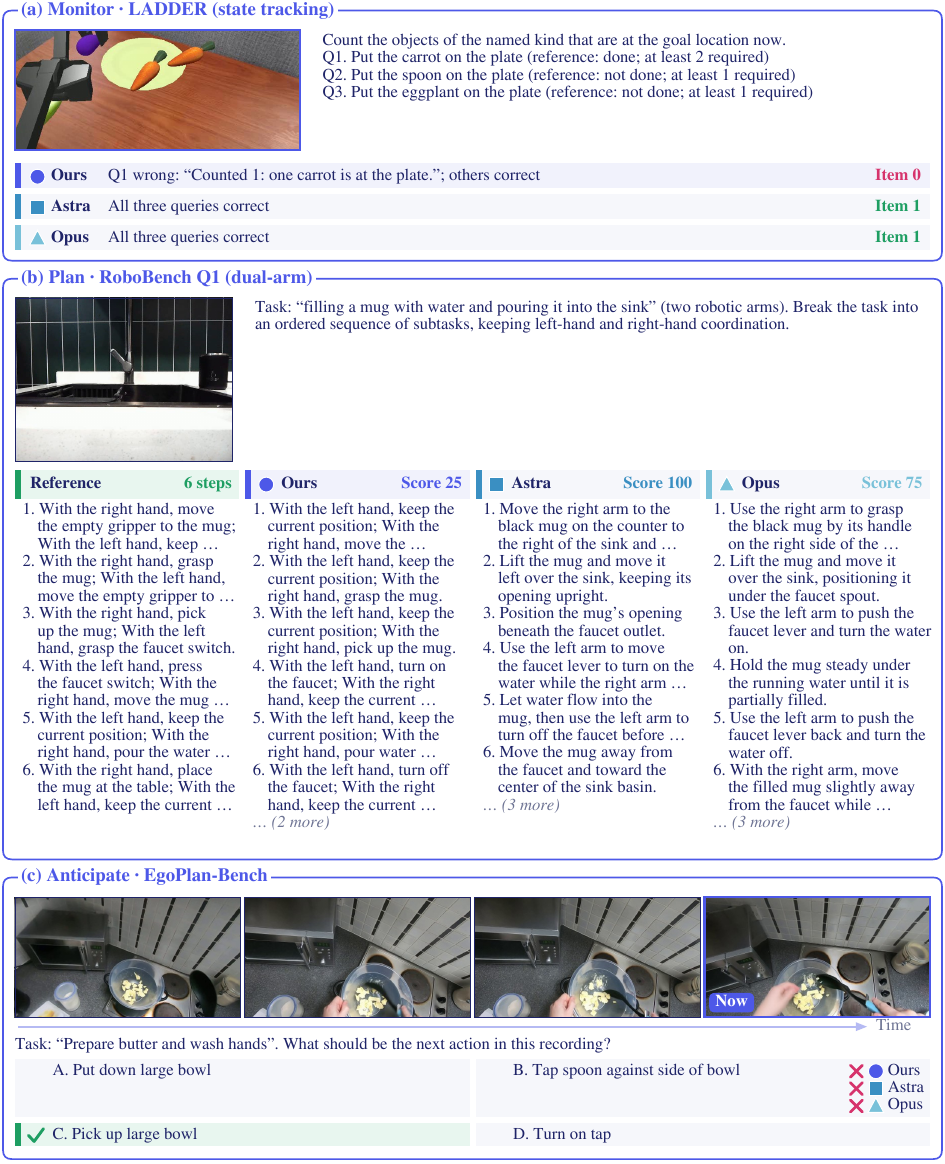}
\caption{\textbf{Further cases} (``Ours'': the task-only model). (a) LADDER state: it answers two of three state queries about a scene correctly but counts one carrot at the plate where two are required, so the scene is scored wrong. (b) RoboBench Q1, dual-arm: it moves one arm at a time, the other idle in almost every step; the reference coordinates the two. (c) EgoPlan-Bench: all three models choose the same wrong option.}
\label{fig:case_more}
\end{figure*}

\medskip
\noindent\textbf{Cases.}\quad
Figures~\ref{fig:cases}, \ref{fig:case_limits_more} and~\ref{fig:case_more} show answers of the task-only model (Section~\ref{sec:data}), marked ``Ours''. Figure~\ref{fig:cases} shows four items on which it is correct and both closed models are wrong, chosen so that the frames show the reference. In (a) and (b), every closed-model answer is a legitimate step of the task, issued from a position the execution has not reached. In (c), both closed models place the failure in the phase in which it becomes visible, one phase after its cause. In (d), they take motion toward a target for arrival. Figure~\ref{fig:case_limit} shows a case in which \ours is wrong, and Figures~\ref{fig:case_limits_more} and~\ref{fig:case_more} cases in which the task-only model is wrong.

\begin{table}[!htb]
\centering
\caption{Datasets of each kind of recording in the training collection (Section~\ref{sec:data}).}
\label{tab:datasets}
\footnotesize
\setlength{\tabcolsep}{4pt}
\begin{tabular}{@{}>{\raggedright\arraybackslash}p{3.2cm}>{\raggedright\arraybackslash}p{7.4cm}>{\raggedright\arraybackslash}p{4.6cm}@{}}
\toprule
Recording & Datasets & Annotations used \\
\midrule
Egocentric human video & EPIC-KITCHENS-100~\cite{damen2020epic}, Ego4D~\cite{grauman2021ego4d}, HoloAssist~\cite{wang2023holoassist}; head-camera recordings annotated by us (Section~\ref{sec:data}) & Time-stamped steps or narrations \\
Third-person robot episodes & DROID~\cite{khazatsky2024droid}, BridgeData~V2~\cite{walke2023bridgedata}, Open X-Embodiment~\cite{oxe2023}, AgiBot World~\cite{bu2025agibot}; failure datasets ViFailback~\cite{zeng2025vifailback} and RoboFAC~\cite{ye2025robofac} & Instructions, trajectories, sub-task lists with boundaries, task labels, failure and success labels \\
Simulation & RoboCasa~\cite{nasiriany2024robocasa} & Exact step boundaries and known object states \\
General-purpose replay & Cambrian~\cite{cambrian} & Replayed without conversion \\
\bottomrule
\end{tabular}
\end{table}

\begin{table}[!htb]
\centering
\caption{Construction of capability questions from annotated recordings (Section~\ref{sec:data}). Every question is emitted only when its answer can be read off the listed annotation, which is kept with the question--answer pair.}
\label{tab:construction}
\footnotesize
\setlength{\tabcolsep}{4pt}
\begin{tabular}{p{1.6cm}p{4.0cm}p{6.4cm}p{2.8cm}}
\toprule
Capability & Annotations used & Question types & Answer form \\
\midrule
Plan & Instruction; Ordered sub-task list of the episode & Decomposition (write the sub-task list from the instruction and first frames); Plan verification (is a candidate plan with a reordered, missing or inserted step valid); Plan continuation (steps remaining after a given moment) & Step list; Yes/No \\
\midrule
Anticipate & Time-stamped steps or narrations & Next step given the frames up to a moment, with task steps as distractors; Free-text next sub-task on robot episodes & Letter; Sub-task sentence \\
\midrule
Monitor & Time-stamped steps; Object states (simulation, state annotations) & Last completed step; Step in progress; Is a named step done; Order of a set of steps; Progress comparison between two clips; State of an individual object & Letter; Yes/No; State label \\
\midrule
Diagnose & Failure labels (type, step or moment, affected state); Successful episodes as negatives & Did the execution fail; Which failure type; At which step or moment; Which object state is wrong & Yes/No; Letter; Location \\
\midrule
Interpret & Task labels; Gripper or hand state & Which task is being performed (choice or words); What has happened so far; Is the gripper holding an object & Letter; Task name; Description; Yes/No \\
\bottomrule
\end{tabular}
\end{table}

\begin{table}[!htb]
\centering
\caption{Composition of the training collection by capability and source domain, in thousands of question--answer pairs (total 3{,}678 thousand).}
\label{tab:comp_matrix}
\footnotesize
\setlength{\tabcolsep}{4pt}
\begin{tabular}{lrrrrrrr}
\toprule
Source & Plan & Anticipate & Monitor & Diagnose & Interpret & General & Total \\
\midrule
Third-person robot (real) & 198.9 & 595.3 & 930.1 & 124.7 & 326.6 & -- & 2{,}175.6 \\
Egocentric human video & 21.4 & 428.1 & 668.8 & 37.3 & 3.2 & -- & 1{,}158.8 \\
Simulation & 18.9 & 30.9 & 27.9 & -- & -- & -- & 77.7 \\
General-purpose & -- & -- & -- & -- & 78.6 & 187.3 & 265.9 \\
\midrule
Total & 239.3 & 1{,}054.3 & 1{,}626.8 & 162.0 & 408.3 & 187.3 & 3{,}678.0 \\
\bottomrule
\end{tabular}
\end{table}

\begin{table}[!htb]
\centering
\caption{Mapping of the five benchmarks to capabilities at the sub-item level. $n$ is the number of questions under the protocol of Section~\ref{sec:exp:setup}. The four LADDER components are computed on one pool of 3{,}409 items, and RoboFAC is reported as the macro average of four question types. ``Judge'' denotes scoring by a language model against the reference; ``MC'' denotes multiple choice.}
\label{tab:map}
\footnotesize
\setlength{\tabcolsep}{5pt}
\begin{tabular}{llrl}
\toprule
Capability & Benchmark sub-item & $n$ & Answer and scoring \\
\midrule
\multirow{2}{*}{Plan}
 & RoboBench Q1: Task decomposition & 1{,}945 & Free-text plan, judge \\
 & LADDER: Decomposition & 145 & Step list, exact match \\
\midrule
\multirow{6}{*}{Anticipate}
 & EgoPlan-Bench & 3{,}343 & 4-way MC \\
 & RoboBench Q2: Next sub-task & 842 & Free text, judge \\
 & LADDER: Next sub-goal & 560 & Sub-goal, validity \\
 & Cosmos-Reason1: AgiBot & 100 & MC \\
 & Cosmos-Reason1: HoloAssist & 100 & MC \\
 & Cosmos-Reason1: BridgeData V2 & 100 & MC \\
\midrule
\multirow{3}{*}{Monitor}
 & RoboBench Q3: Step completion & 1{,}192 & Yes/No \\
 & LADDER: Verification & 562 & Yes/No, balanced accuracy \\
 & LADDER: State & 2{,}142 & All state queries of a scene correct (562 scenes) \\
\midrule
\multirow{4}{*}{Diagnose}
 & RoboFAC: Failure detection & 1{,}204 & Yes/No \\
 & RoboFAC: Failure identification & 960 & 6-way MC \\
 & RoboFAC: Failure locating & 960 & MC \\
 & Cosmos-Reason1: RoboFail & 100 & MC on sub-task and task success \\
\midrule
\multirow{2}{*}{Interpret}
 & RoboFAC: Task identification & 1{,}204 & Open-ended, judge \\
 & Cosmos-Reason1: RoboVQA & 110 & MC \\
\bottomrule
\end{tabular}
\end{table}

\begin{table}[!htb]
\centering
\caption{Scores on the seventeen sub-items of the five benchmarks, grouped by capability (Table~\ref{tab:map}); the values plotted in Figure~\ref{fig:profile}. $\Delta$ is the difference between \ours and the better of the two closed models. Best of \ours and the two closed models in \textbf{bold}.}
\label{tab:subitems}
\footnotesize
\setlength{\tabcolsep}{6pt}
\begin{tabular}{llccccr}
\toprule
Capability & Sub-item & \oursTwoLine & \makecell{GPT-6\\Astra} & \makecell{Claude\\Opus 5.5} & Base & $\Delta$ \\
\midrule
\multirow{2}{*}{Plan}
 & RoboBench Q1: task decomposition & 67.7 & 74.5 & \textbf{77.8} & 64.9 & $-10.1$ \\
 & LADDER: decomposition & 93.1 & 86.9 & \textbf{93.8} & 92.4 & $-0.7$ \\
\midrule
\multirow{6}{*}{Anticipate}
 & EgoPlan-Bench & 74.5 & \textbf{75.9} & 70.1 & 48.8 & $-1.4$ \\
 & RoboBench Q2: next sub-task & \textbf{52.1} & 34.7 & 39.0 & 22.3 & $+13.1$ \\
 & LADDER: next sub-goal & 76.8 & \textbf{84.6} & 79.5 & 64.3 & $-7.8$ \\
 & Cosmos-Reason1: AgiBot & \textbf{94.3} & 69.0 & 69.5 & 47.7 & $+24.8$ \\
 & Cosmos-Reason1: HoloAssist & \textbf{88.0} & 84.0 & 69.5 & 73.7 & $+4.0$ \\
 & Cosmos-Reason1: BridgeData V2 & 65.3 & 67.5 & \textbf{72.5} & 32.0 & $-7.2$ \\
\midrule
\multirow{3}{*}{Monitor}
 & RoboBench Q3: step completion & \textbf{77.8} & 74.1 & 75.4 & 63.9 & $+2.4$ \\
 & LADDER: verification & \textbf{78.6} & 78.2 & 76.5 & 74.1 & $+0.4$ \\
 & LADDER: state & 67.6 & 53.2 & \textbf{68.3} & 44.1 & $-0.7$ \\
\midrule
\multirow{4}{*}{Diagnose}
 & RoboFAC: failure detection & 83.3 & 79.8 & \textbf{86.0} & 42.8 & $-2.7$ \\
 & RoboFAC: failure identification & \textbf{58.6} & 48.4 & 52.2 & 35.9 & $+6.4$ \\
 & RoboFAC: failure locating & \textbf{48.5} & 38.9 & 37.6 & 38.0 & $+9.6$ \\
 & Cosmos-Reason1: RoboFail & 68.0 & 78.5 & \textbf{86.5} & 63.0 & $-18.5$ \\
\midrule
\multirow{2}{*}{Interpret}
 & RoboFAC: task identification & \textbf{86.2} & 45.3 & 53.9 & 26.8 & $+32.3$ \\
 & Cosmos-Reason1: RoboVQA & \textbf{100.0} & 87.7 & 83.6 & 90.9 & $+12.3$ \\
\bottomrule
\end{tabular}
\end{table}

\begin{table}[!htb]
\centering
\caption{RoboBench Q1 (task decomposition) by embodiment, object attribute, camera views and reference plan length for \ours, \astra and \opus.}
\label{tab:rb_q1}
\footnotesize
\setlength{\tabcolsep}{5pt}
\begin{tabular}{llrccc}
\toprule
Group & Subset & $n$ & \oursTwoLine & GPT-6 Astra & Claude Opus 5.5 \\
\midrule
\multirow{4}{*}{Embodiment}
 & Single arm & 235 & 85.7 & 84.5 & 88.3 \\
 & Dual arm & 157 & 36.5 & 64.4 & 64.9 \\
 & Human & 118 & 79.5 & 79.2 & 84.9 \\
 & Mobile manipulation & 50 & 45.6 & 66.9 & 72.7 \\
\midrule
\multirow{3}{*}{Object attribute}
 & Shape & 126 & 83.1 & 69.7 & 73.4 \\
 & Colour & 123 & 88.6 & 81.6 & 81.5 \\
 & Size & 92 & 87.3 & 74.8 & 70.9 \\
\midrule
\multirow{2}{*}{Views}
 & Single view & 80 & 49.1 & 62.6 & 66.2 \\
 & Multiple views & 82 & 52.1 & 64.0 & 64.1 \\
\midrule
Objects & Counted set of objects & 117 & 36.8 & 54.1 & 58.7 \\
\midrule
\multirow{4}{*}{Reference plan length}
 & 1--3 steps & 572 & 78.4 & 80.1 & 82.4 \\
 & 4--6 steps & 700 & 72.4 & 79.0 & 82.9 \\
 & 7--11 steps & 443 & 62.2 & 72.0 & 76.7 \\
 & 12 or more steps & 230 & 37.6 & 51.7 & 53.0 \\
\bottomrule
\end{tabular}
\end{table}

\begin{table}[!htb]
\centering
\caption{EgoPlan-Bench by source, by whether the frames show progress toward the goal, and by the length of the correct option relative to the other options.}
\label{tab:egoplan_subsets}
\footnotesize
\setlength{\tabcolsep}{5pt}
\begin{tabular}{llrccc}
\toprule
Group & Subset & $n$ & \oursTwoLine & GPT-6 Astra & Claude Opus 5.5 \\
\midrule
\multirow{2}{*}{Source}
 & EPIC-KITCHENS & 2{,}420 & 74.7 & 75.2 & 70.1 \\
 & Ego4D & 923 & 74.2 & 77.8 & 70.5 \\
\midrule
\multirow{2}{*}{Progress frames}
 & Present & 2{,}841 & 75.0 & 76.9 & 71.3 \\
 & Absent (answer is the first step) & 502 & 71.9 & 69.9 & 64.1 \\
\midrule
\multirow{3}{*}{Correct option is}
 & the shortest & 739 & 69.4 & 73.3 & 66.6 \\
 & of middle length & 1{,}755 & 75.4 & 74.7 & 70.3 \\
 & the longest & 849 & 77.3 & 80.6 & 73.1 \\
\bottomrule
\end{tabular}
\end{table}

\clearpage

\end{document}